\documentclass[12pt]{colt2018}

\usepackage{microtype} 
\usepackage{algorithm}
\usepackage{algpseudocode}
\usepackage{parskip}

\usepackage{mathtools}
\usepackage{amsmath}
\usepackage{bbm}
\usepackage{amsfonts}
\usepackage{amssymb}

\usepackage{graphicx}
\usepackage{latexsym}
\usepackage[mathscr]{euscript}
\usepackage{multirow}
\usepackage{times}

\usepackage{MnSymbol}
\usepackage{xpatch}

\def\Rset{\mathbb{R}}

\def\sfD{\mathsf{D}}

\let\Pr\relax 
\DeclareMathOperator*{\Pr}{\mathbb{P}}

\DeclareMathOperator*{\E}{\mathbb E}

\DeclareMathOperator*{\argmin}{argmin}

\DeclareMathOperator*{\conv}{conv}
\DeclareMathOperator*{\var}{Var}

\newtheorem*{rep@theorem}{\rep@title}
\newcommand{\newreptheorem}[2]{%
\newenvironment{rep#1}[1]{%
 \def\rep@title{#2 \ref{##1}}%
 \begin{rep@theorem}}%
 {\end{rep@theorem}}}

\newreptheorem{theorem}{Theorem}
\newreptheorem{lemma}{Lemma}
\newreptheorem{corollary}{Corollary}
\newreptheorem{proposition}{Proposition}

\newtheorem{properties}{Properties}

\newcommand{\cH}{\mathcal{H}}

\newcommand{\cL}{\mathcal{L}}

\newcommand{\cX}{\mathcal{X}}

\newcommand{\sD}{{\mathscr D}}
\newcommand{\sG}{{\mathscr G}}
\newcommand{\sH}{{\mathscr H}}

\newcommand{\sL}{{\mathscr L}}

\newcommand{\sU}{{\mathscr U}}
\newcommand{\sW}{{\mathscr W}}
\newcommand{\sX}{{\mathscr X}}
\newcommand{\sY}{{\mathscr Y}}

\newcommand{\ba}{{\mathbf a}}

\newcommand{\bm}{{\mathbf m}}

\newcommand{\bx}{{\mathbf x}}
\newcommand{\by}{{\mathbf y}}

\renewcommand{\L}{\mathsf{L}}

\newcommand{\R}{{\mathfrak R}}
\newcommand{\s}{{\mathfrak s}}

\newcommand{\bsigma}{{\boldsymbol \sigma}}
\newcommand{\para}[1][]{\, #1\| \,}

\newcommand{\h}{\widehat}

\newcommand{\ov}{\overline}

\newcommand{\e}{\epsilon}
\renewcommand{\set}[2][]{#1 \{ #2 #1 \} }
\newcommand{\ignore}[1]{}

\hypersetup{
  colorlinks   = true,
  urlcolor     = blue,
  linkcolor    = blue,
  citecolor   = blue
}

\title{Agnostic Federated Learning}

\coltauthor{\Name{Mehryar Mohri}\Email{mohri@google.com}\\
\addr Google Research and 
Courant Institute of Mathematical Sciences, New York
\AND
\Name{Gary Sivek}\Email{gsivek@google.com}\\
\addr Google Research, New York
\AND
\Name{Ananda Theertha Suresh}\Email{theertha@google.com}\\
\addr Google Research, New York
}

\newcommand{\icml}[1]{}
\newcommand{\arxiv}[1]{#1}
\begin{document}

\maketitle

\begin{abstract}
  A key learning scenario in large-scale applications is that of
  \emph{federated learning}, where a centralized model is trained
  based on data originating from a large number of clients. We argue
  that, with the existing training and inference, federated models can
  be biased towards different clients. Instead, we propose a new
  framework of \emph{agnostic federated learning}, where the
  centralized model is optimized for any target distribution formed by
  a mixture of the client distributions. We further show that this
  framework naturally yields a notion of fairness. We present
  data-dependent Rademacher complexity guarantees for learning with
  this objective, which guide the definition of an algorithm for
  agnostic federated learning. We also give a fast stochastic
  optimization algorithm for solving the corresponding optimization
  problem, for which we prove convergence bounds, assuming a convex
  loss function and hypothesis set. We further empirically demonstrate
  the benefits of our approach in several datasets. Beyond federated
  learning, our framework and algorithm can be of interest to other
  learning scenarios such as cloud computing, domain adaptation,
  drifting, and other contexts where the training and test
  distributions do not coincide.
\end{abstract}

\section{Motivation}

A key learning scenario in large-scale applications is that of
\emph{federated learning}. In that scenario, a centralized model is
trained based on data originating from a large number of clients,
which may be mobile phones, other mobile devices, or sensors
\citep*{konevcny2016federated,konecny2016federated2}.  The training
data typically remains distributed over the clients, each with
possibly unreliable or relatively slow network connections.

Federated learning raises several types of issues and has been the
topic of multiple research efforts. These include systems, networking
and communication bottleneck problems due to frequent exchanges
between the central server and the clients \icml{
  \cite{McMahanMooreRamageHampsonAguera2017}}.\arxiv{ To deal with
  such problems, \cite{McMahanMooreRamageHampsonAguera2017} suggested
  an averaging technique that consists of transmitting the central
  model to a subset of clients, training it with the data locally
  available, and averaging the local updates.
  \cite{SmithChiangSanjabiTalwalkar2017} proposed to further leverage
  the relationship between clients, assumed to be known, and cast the
  problem as an instance of multi-task learning to derive local client
  models benefiting from other similar ones.

The optimization task in federated learning, which is a principal
problem in this scenario, has also been the topic of multiple research
work. That includes } \icml{Other research efforts include }the design
of more efficient communication strategies
\citep*{konevcny2016federated,konecny2016federated2,
  suresh2017distributed}, devising efficient distributed optimization
methods benefiting from differential privacy guarantees
\citep*{AgarwalSureshYuKumarMcMahan2018}, as well as recent guarantees
for parallel stochastic optimization with a dependency graph
\citep*{WoodworthWangSmithMcMahanSrebro2018}.

Another key problem in federated learning which appears more generally
in distributed machine learning and other learning setups is that of
\emph{fairness}.  In many instances in practice, the resulting
learning models may be biased or unfair: they may discriminate against
some protected groups \citep*{Bickel398,hardt2016equality}. As a simple
example, a regression algorithm predicting a person's salary could be
using that person's gender. This is a key problem in modern machine
learning that does not seem to have been specifically studied in the
context of federated learning.

While many problems related to federated learning have been
extensively studied, the key objective of learning in that context
seems not to have been carefully examined.  We are also not aware of
statistical guarantees derived for learning in this scenario. A
crucial reason for such questions to emerge in this context is that
the target distribution for which the centralized model is learned is
unspecified. Which expected loss is federated learning seeking to
minimize? 
Most centralized models for standard federated learning are
trained on the aggregate training sample obtained from the subsamples
drawn from the clients. Thus, if we denote by $\sD_k$ the distribution
associated to client $k$, $m_k$ the size of the sample available from
that client and $m$ the total sample size, intrinsically, the
centralized model is trained to minimize the loss with respect to the
\emph{uniform distribution}
\arxiv{\[}\icml{$}
\ov \sU = \textstyle \sum_{k = 1}^p \frac{m_k}{m} \sD_k.
\arxiv{\]}\icml{$}
 But why
should $\ov \sU$ be the target distribution of the learning model? Is
$\ov \sU$ the distribution that we expect to observe at test time?
What guarantees can be derived for the deployed system?

Notice that, in practice, in federated learning, the probability that
an individual data source participates in training depends on various
factors such as whether the mobile device is connected to the internet
or whether it is being charged. Thus, the training data may not truly
reflect the usage of the learned model in inference. Additionally,
these uncertainties may also affect the size of the sample $m_k$
acquired from each client, which directly affects the definition of
$\ov \sU$.

We argue that in many common instances, the uniform distribution is
not the natural objective distribution and that seeking to minimize
the expected loss with respect to the specific distribution $\ov \sU$
is \emph{risky}. This is because the target distribution may be in
general quite different from $\ov \sU$.  In many cases, that can
result in a suboptimal or even a detrimental performance. For example,
imagine a plausible scenario of federated learning where the learner
has access to a large population of expensive mobile phones, which are
most commonly adopted by software engineers or other technical users
(say $70\%$) than other users ($30\%$), and a small population of
other mobile phones less used by non-technical users ($5\%$) and
significantly more often by other users ($95\%$). The centralized
model would then be essentially based on the uniform distribution
based on the expensive clients. But, clearly, such a model would not be
adapted to the wide general target domain formed by the majority of
phones with a $5\%\mathord-95\%$ population of general versus
technical users. Many other realistic examples of this type can help
illustrate the learning problem resulting from a mismatch between the
target distribution and $\ov \sU$. In fact, it is not clear why
minimizing the expected loss with respect to $\ov \sU$ could be
beneficial for the clients, whose distributions are $\sD_k$s.

Thus, we put forward a new framework of \emph{agnostic federated
  learning} (AFL), where the centralized model is optimized for any possible
target distribution formed by a mixture of the client distributions.
Instead of optimizing the centralized model for a specific
distribution, with the high risk of a mismatch with the target, we
define an agnostic and more risk-averse objective. We show that,
for some target mixture distributions, the cross-entropy loss of the
hypothesis obtained by minimization with respect to the uniform
distribution $\ov \sU$ can be worse, by a constant additive term,
than that of the hypothesis obtained in AFL,
even if the learner has access to an infinite sample size
(Section~\ref{sec:cmp}).

We further show that our AFL framework naturally yields a notion of
fairness, which we refer to as \emph{good-intent fairness}
(Section~\ref{sec:fairness}).  Indeed, the predictor solution of the
optimization problem for our AFL framework treats all protected
categories similarly.
Beyond federated learning, our framework and solution also cover
related problems in cloud-based learning services, where customers may
not have any training data at their disposal or may not be willing to
share that data with the cloud. In that case too, the server needs to
train a model without access to the training data. Our framework and
algorithm can also be of interest to other learning scenarios such as
domain adaptation, drifting, and other contexts where the training and
test distributions do not coincide. \icml{In
  Appendix~\ref{sec:related}, we give an extensive discussion of
  related work, including connections with the broad literature of
  domain adaptation.}

The rest of the paper is organized as follows.  \arxiv{ In
  Section~\ref{sec:related}, we give an extensive discussion of
  related work, including connections with the broad literature of
  domain adaptation.}  In Section~\ref{sec:scenario}, we give a formal
description of \arxiv{the learning scenario of federated learning and
  the formulation of the problem as} AFL. Next, we give a detailed
theoretical analysis of learning in the AFL framework\arxiv{,
  including data-dependent Rademacher complexity generalization
  bounds} (Section~\ref{sec:theory}). \arxiv{These bounds lead to a
  natural learning algorithm with a regularization term based on a
  \emph{skewness term} that we define (Section~\ref{sec:algorithm}).}
We also present an efficient convex optimization algorithm for solving
the optimization problem defining our algorithm
(Section~\ref{sec:optimization}). \arxiv{Our algorithm is a stochastic
  gradient-descent solution for minimax problems, for which we give a
  detailed analysis, including the proof of convergence in terms of
  the variances of the stochastic gradients.} In
Section~\ref{sec:experiments}, we present a series of experiments
comparing our AFL algorithm and solution with existing federated
learning solutions. In
\icml{Appendix}\arxiv{Section}~\ref{sec:extensions}, we discuss
several extensions of AFL\icml{ such as how to incorporate priors,
  personalization and how to choose domains}.

\arxiv{\section{Related work}
\label{sec:related}

Here, we briefly discuss several learning scenarios and work related
to our study of federated learning.

The problem of federated learning is closely related to other learning
scenarios where there is a mismatch between the source distribution
and the target distribution. This includes the problem of
\emph{transfer learning} or \emph{domain adaptation} from a single
source to a known target domain
\citep*{BenDavidBlitzerCrammerPereira2006,MansourMohriRostamizadeh2009Bis,CortesMohri2014,CortesMohriMunoz2015},
either through unsupervised adaptation techniques \citep{gong_cvpr12,
  long_icml15, ganin_icml15, tzeng_iccv15}, or via lightly supervised
ones (some amount of labeled data from the target domain)
\citep{saenko_eccv10, yang_acmm07, hoffman_iclr13, rcnn}. This also
includes previous applications in natural language processing
\citep{Dredze07Frustratingly,Blitzer07Biographies,jiang-zhai07,
  raju2018contextual}, speech recognition
\citep{Legetter&Woodlang,Gauvain&Lee,DellaPietra,
  Rosenfeld96,jelinek,roark03supervised}, and computer vision
\citep{martinez}

A problem more closely related to that of federated learning is that
of \emph{multiple-source adaptation}, first formalized and
analyzed theoretically by
\citet*{MansourMohriRostamizadeh2009a,MansourMohriRostamizadeh2009} and
later studied for various applications such as object recognition
\citep{hoffman_eccv12, gong_icml13, gong_nips13}.  Recently,
\cite{zhang2015multi} studied a causal formulation of this problem for
a classification scenario, using the same combination rules as
\citet{MansourMohriRostamizadeh2009a,MansourMohriRostamizadeh2009}.
The problem of \emph{domain generalization}
\citep{pan_tkda2010,MuandetBalduzziScholkopf2013,xu_eccv14}, where
knowledge from an arbitrary number of related domains is combined to
perform well on a previously unseen domain is very closely related to
that of federated learning, though the assumptions about the
information available to the learner and the availability of unlabeled
data may differ.

In the multiple-source adaptation problem studied by
\citet*{MansourMohriRostamizadeh2009a,MansourMohriRostamizadeh2009} and
\citet*{HoffmanMohriZhang2018}, each domain $k$ is defined by the
corresponding distribution $\sD_k$ and the learner has only access to
a predictor $h_k$ for each domain and no access to labeled training
data drawn from these domains. The authors show that it is possible
to define a predictor $h$ whose expected loss $\sL_\sD(h)$ with
respect to any distribution $\sD$ that is a mixture of the source
domains $\sD_k$ is at most the maximum expected loss of the source
predictors: $\max_{k} L_{\sD_k}(h_{\sD_k})$. They also provide an 
algorithm for determining $h$.

Our learning scenario differs from the one adopted in that work since
we assume access to labeled training data from each domain
$\sD_k$. Furthermore, the predictor determined by the algorithm of
\citet*{HoffmanMohriZhang2018} belongs to a specific hypothesis set
$\sH'$, which is that of distribution weighted combinations of the
domain predictors $h_k$, while, in our setup, the objective is to
determine the best predictor in some global hypothesis set $\sH$,
which may include $\sH'$ as a subset, and which is not depending
on some domain-specific predictors.

Our optimization solution also differs from the work of
\citet{farnia2016minimax} and \citet{lee2017minimax} on local minimax
results, where samples are drawn from a single source $\sD$, and where
the generalization error is minimized over a set of locally ambiguous
distributions $\h \sD$, where $\h \sD$ is the empirical
distribution. The authors propose this metric for statistical
robustness.  In our work, we obtain samples from $p$ unknown
distributions, and the set of distributions $D_\lambda$ over which we
optimize the expected loss is fixed and independent of
samples. Furthermore, the source distributions can differ arbitrarily
and need not be close to each other. In reverse, we note that
our stochastic algorithm can be used to minimize the loss
functions proposed in \citep{farnia2016minimax,lee2017minimax}.

}
\section{Learning scenario}
\label{sec:scenario}

In this section, we introduce the learning scenario of agnostic
federated learning we consider. Next, we first argue that the uniform
solution commonly adopted in standard federated learning may not be an
adequate solution, thereby further justifying our agnostic model.
Second, we show the benefit of our model in fairness learning.

We start with some general notation and definitions used throughout
the paper. Let $\sX$ denote the input space and $\sY$ the output
space. We will primarily discuss a multi-class classification problem
where $\sY$ is a finite set of classes, but much of our results can be
extended straightforwardly to regression and other problems.
The hypotheses we consider are of the form
$h\colon \sX \to \Delta_\sY$, where $\Delta_\sY$ stands for the
simplex over $\sY$. Thus, $h(x)$ is a probability distribution over
the classes or categories that can be assigned to $x \in \sX$. We will
denote by $\sH$ a family of such hypotheses $h$.  We also denote by
$\ell$ a loss function defined over $\Delta_\sY \times \sY$ and taking
non-negative values. The loss of $h \in \sH$ for a labeled sample
$(x, y) \in \sX \times \sY$ is given by $\ell(h(x), y)$. One key
example in applications is the cross-entropy loss, which is defined as
follows: $\ell(h(x), y) = -\log (\Pr_{y' \sim h(x)}[y' = y])$.  We
will denote by $\sL_\sD(h)$ the expected loss of a hypothesis $h$ with
respect to a distribution $\sD$ over $\sX \times \sY$:
\[
\sL_\sD(h) = \E_{(x, y) \sim \sD} [\ell(h(x), y)],
\]
and by $h_\sD$ its minimizer:
$h_\sD = \argmin_{h \in \sH} \sL_\sD(h)$. 

\subsection{Agnostic federated learning}
\label{sec:afl}

We consider a learning scenario where the learner receives $p$ samples
$S_1, \ldots, S_p$, with each
$S_k = ((x_{k, 1}, y_{k, 1}), \ldots, (x_{k, m_k}, y_{k, m_k})) \in
(\sX \times \sY)^{m_k}$ of size $m_k$ drawn i.i.d.\ from a different
domain or distribution $\sD_k$.  The learner's objective is to
determine a hypothesis $h \in \sH$ that performs well on some target
distribution.  We will also denote by $\h \sD_k$ the empirical
distribution associated to sample $S_k$ of size $m$ drawn from
$\sD^m$.

\begin{figure}[t]
\centering
\arxiv{\includegraphics[scale=.5]{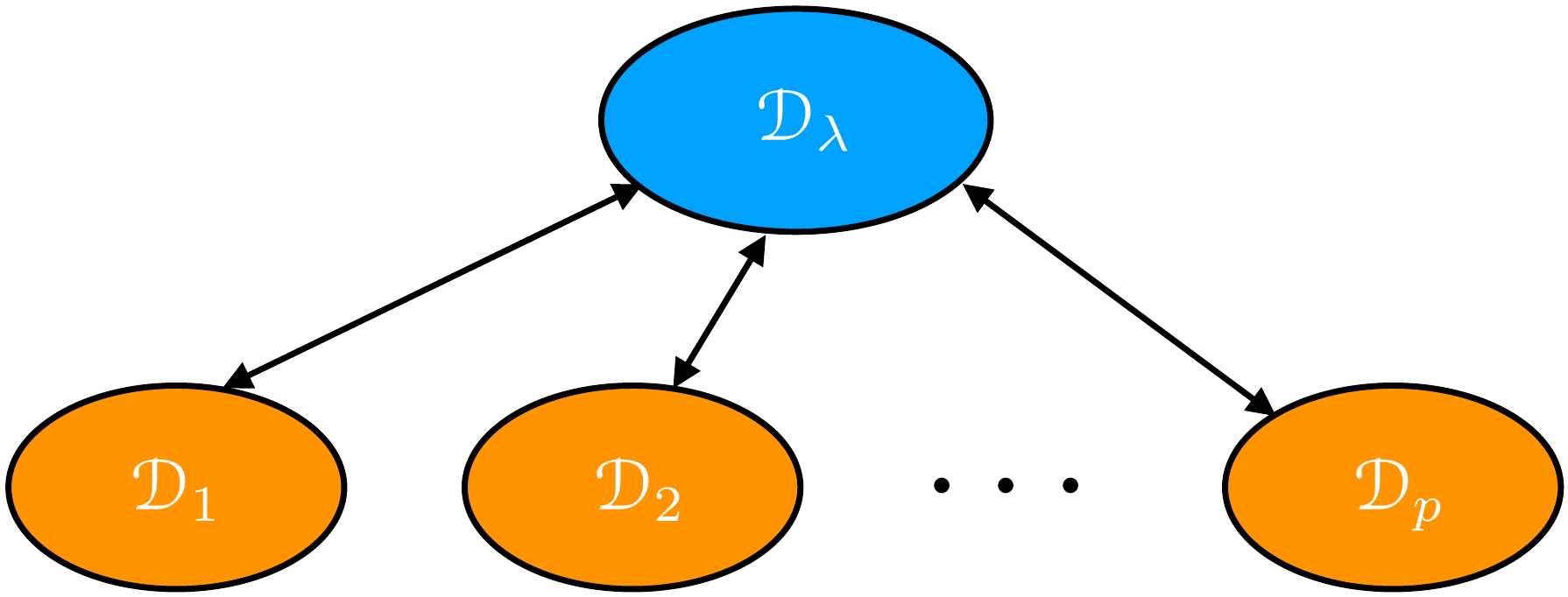}}
\icml{\includegraphics[scale=.4]{fig1.pdf}}
\caption{Illustration of the agnostic federated learning scenario.}
\label{fig:agnostic}
\end{figure}

This scenario coincides with that of \emph{federated learning} where
training is done with the \emph{uniform distribution} over the union
of all samples $S_k$, that is
$\h \sU = \sum_{k = 1}^p \frac{m_k}{\sum_{k = 1}^p m_k} \h \sD_k$, and
where the underlying assumption is that the target distribution is
$\ov \sU = \sum_{k = 1}^p \frac{m_k}{\sum_{k = 1}^p m_k} \sD_k$. We
will not adopt that assumption since it is rather restrictive and since,
as discussed later, it can lead to solutions that are disadvantageous
to domain users.  Instead, we will consider an \emph{agnostic
  federated learning} (AFL) scenario where the target distribution can
be modeled as an unknown mixture of the distributions $\sD_k$,
$k = 1, \ldots, p$, that is
$\sD_\lambda = \sum_{k = 1}^p \lambda_k \sD_k$ for some
$\lambda \in \Delta_p$. Since the mixture weight $\lambda$ is unknown,
here, the learner must come up with a solution that is favorable for
any $\lambda$ in the simplex, or any $\lambda$ in a subset
$\Lambda \subseteq \Delta_p$.
Thus, we define the \emph{agnostic loss} (or \emph{agnostic risk})
$\sL_{\sD_\Lambda}(h)$ associated to a predictor $h \in \sH$ as
\begin{equation}
\label{eq:AgnosticRisk}
\sL_{\sD_\Lambda}(h) = \max_{\lambda \in \Lambda} \sL_{\sD_\lambda}(h).
\end{equation}
We will extend our previous definitions and denote by
$h_{\sD_\Lambda}$ the minimizer of this loss:
\arxiv{\[}\icml{$}
h_{\sD_\Lambda}= \argmin_{h \in \sH} \sL_{\sD_\Lambda}(h).
\arxiv{\]}\icml{$}

In practice, the learner has access to the distributions $\sD_k$ only
via the finite samples $S_k$. Thus, for any $\lambda \in \Delta_p$,
instead of the mixture $\sD_\lambda$, only the $\lambda$-mixture of
empirical distributions,
$\ov \sD_\lambda = \sum_{k=1}^p \lambda_k \h \sD_k$, is
accessible.\footnote{Note, $\ov \sD_\lambda$ is distinct from an
  empirical distribution $\h \sD_\lambda$ which would be based on a
  sample drawn from $\sD_\lambda$. $\ov \sD_\lambda$ is based on
  samples drawn from $\sD_k$s.} This leads to the definition of
$\sL_{\ov \sD_{\Lambda}}(h)$, the \emph{agnostic empirical loss} of a
hypothesis $h \in \sH$ for a subset of the simplex $\Lambda$:
\begin{equation*}
\sL_{\ov \sD_{\Lambda}}(h) = \max_{\lambda \in \Lambda}
\sL_{\ov \sD_{\lambda}}(h).
\end{equation*}
We will denote by $h_{\ov \sD_{\Lambda}}$ the minimizer of this loss:
$h_{\ov \sD_{\Lambda}} = \argmin_{h \in \cH} \sL_{ \ov
  \sD_{\Lambda}}(h)$. In the next section, we will present
generalization bounds relating the expected and empirical agnostic
losses $\sL_{\sD_{\Lambda}}(h)$ and $\sL_{\ov \sD_{\Lambda}}(h)$ for
all $h \in \sH$. 

Notice that the domains $\sD_k$ discussed thus far need not coincide
with the clients. In fact, when the number of clients is very large
and $\Lambda$ is the full simplex, $\Lambda = \Delta_p$, it is
typically preferable to consider instead domains defined by clusters
of clients, as discussed in
\arxiv{Section}\icml{Appendix}~\ref{sec:extensions}.  On the other
hand, if $p$ is small or $\Lambda$ more restrictive, then the model
may not perform well on certain domains of interest. We mitigate the
effect of large $p$ values using a suitable regularization term
derived from our theory.

\subsection{Comparison with federated learning}
\label{sec:cmp}

Here, we further argue that the uniform solution $h_{\ov \sU}$
commonly adopted in federated learning may not provide a satisfactory
performance compared with a solution of the agnostic problem. This
further motivates our AFL model.

As already discussed, since the target distribution is unknown, the
natural method for the learner is to select a hypothesis minimizing
the agnostic loss $\sL_{\sD_\Lambda}$. Is the predictor minimizing the
agnostic loss coinciding with the solution $h_{\h \sU}$ of standard
federated learning? How poor can the performance of the standard
federated learning be?  We first show that the loss of $h_{\h \sU}$
can be higher than that of the optimal loss achieved by
$h_{\sD_\Lambda}$ by a constant loss, even if the number of samples
tends to infinity, that is even if the learner has access to the
distributions $\sD_k$ and uses the predictor $h_{\ov \sU}$. \arxiv{Similar
results are known for universal compression, where the goal is to
compress a sequence of random variables without knowledge of the
generating distribution \citep{grunwald2007minimum}.}
\begin{proposition}\icml{[Appendix~\ref{app:lower_bound}]}
\label{pro:lower_bound}
  Let $\ell$ be the cross-entropy loss. Then, there exist $\Lambda$,
  $\sH$, and $\sD_k$, $k \in [p]$, such that the following inequality
  holds:
\[
\sL_{\sD_{\Lambda}}(h_{\ov \sU}) \geq \sL_{\sD_{\Lambda}}(h_{\sD_\Lambda}) + \log
\frac{2}{\sqrt{3}}.
\]
\end{proposition}
\arxiv{\begin{proof}
  Consider the following two distributions with support reduced to a
  single element $x \in \sX$ and two classes $\sY = \set{0, 1}$:
  $\sD_1(x, 0) = 0$, $\sD_2(x, 1) = 1$, $\sD_2(x, 0) = \frac{1}{2}$,
  and $\sD_2(x, 1) = \frac{1}{2}$.  Let
  $\Lambda = \set{\delta_1, \delta_2}$, where $\delta_k$, $k = 1, 2$,
  denotes the Dirac measure on index $k$.  We will consider the case
  where the sample sizes $m_k$ are all equal, that is
  $h_{\ov \sU} = \frac{1}{2} (\sD_1 + \sD_2)$.  Let $p_0$ denote the
  probability that $h$ assigns to class $0$ and $p_1$ the one
  it assigns to class $1$. Then, the cross-entropy loss of a predictor
  $h$ can be expressed as follows:
\begin{align*}
  \sL_{\ov \sU}(h) 
= \E_{(x, y) \sim \ov \sU}\big[-\log p_y \big]
& = \frac{1}{4} \log \frac{1}{p_0} + \frac{1}{2} \log \frac{1}{p_1} + \frac{1}{4} \log \frac{1}{p_1}\\
& = \frac{1}{4} \log \frac{1}{p_0} + \frac{3}{4} \log \frac{1}{p_1} \\
& = \sfD \big(\big(\tfrac{1}{4}, \tfrac{3}{4} \big) \para (p_0, p_1)
  \big) + \frac{1}{4} \log \frac{4}{1} +
\frac{3}{4} \log \frac{4}{3} \\
& \geq \frac{1}{4} \log \frac{4}{1} + \frac{3}{4} \log \frac{4}{3} ,
\end{align*}
where the last inequality follows the non-negativity of the relative
entropy. Furthermore, equality is achieved when
$p_0 = 1 - p_1 = \frac{1}{4}$, which defines $h_{\ov \sU}$, the
minimizer of $\sL_{\ov \sU}(h)$. In view of that,
$\cL_{\sD_{\Lambda}}(h_{\ov \sU})$ is given by the following:
\begin{align*}
\cL_{\sD_{\Lambda}}(h_{\ov \sU}) & 
= \max \left( \sL_{\delta_1}(\ov \sU), \sL_{\delta_2}(\ov \sU) \right) \\ 
& = \max \set[\bigg]{\log \frac{4}{3}, 
\frac{1}{2} \log \frac{4}{1} + \frac{1}{2} \log \frac{4}{3} }\\
& = \log \frac{4}{\sqrt{3}}.
\end{align*}
We now compute the loss of $h_{\sD_\Lambda}$:
\begin{align*}
\min_{h \in \sH} \sL_{\sD_{\Lambda}}(h) 
& = \min_{h \in \sH} \max_{k \in [p]} \sL_{\sD_k}(h)\\
& = \min_{(p_0, p_1) \in \Delta_2} \max \set[\Bigg]{\log \frac{1}{p_1}, \frac{1}{2} \log
  \frac{1}{p_0} + \frac{1}{2} \log \frac{1}{p_1} } \\
& = \min_{p_1 \in [0, 1]} \max \set[\Bigg]{\log \frac{1}{p_1}, \log \frac{1}{\sqrt{ p_1 (1 - p_1)}} } \\
& = \log 2,
\end{align*}
since $\frac{1}{2}$ is the solution of the convex optimization in
$p_1$, in view of
$\max \set[\Big]{\frac{1}{p_1}, \frac{1}{\sqrt{ p_1 (1 - p_1)}}} =
\frac{1}{\sqrt{p_1 (1 - p_1)}} \leq \frac{1}{2}$ for
$p_1 > \frac{1}{2}$.

\end{proof}}

\subsection{Good-intent fairness in learning} 
\label{sec:fairness}

\arxiv{Here, we further discuss the relationship between our model
of AFL and fairness in learning.}

Fairness in machine learning has received much attention in recent
past \citep{Bickel398, hardt2016equality}. There is now a broad
literature on the topic with a variety of definitions of the notion of
fairness. In a typical scenario, there is a protected class $c$ among
$p$ classes $c_1, c_2, \ldots, c_p$. While there are many definitions
of fairness, the main objective of a fairness algorithm is to reduce
bias and ensure that the model is fair to all the $p$ protected
categories, under some definition of fairness. The most common reasons
for bias in machine learning algorithms are training data bias and
overfitting bias. We first provide a brief explanation and
illustration for both:
\begin{itemize}

\item the training data is biased: consider the regression task, where
  the goal is to predict the salary of a person based on features such
  as education, location, age, gender. Let gender be the protected
  class. If in the training data, there is a consistent
  discrimination against women irrespective of their education, e.g.,
  their salary is lower, then we can conclude that the training data
  is inherently biased.

\item the training procedure is biased: consider an image
  recognition task where the protected category is race. If the model is
  heavily trained on images based on certain races, then the resulting
  model will be biased because of over-fitting.

\end{itemize}

Our model of AFL can help define a notion of
good-intent fairness, where we reduce the bias in the training
procedure. Furthermore, if training procedure bias exists, it
naturally highlights it.

Suppose we are interested in a classification problem and there is a
protected feature class $c$, which can be one of $p$ values
$c_1, c_2, \ldots, c_p$. Then, we define $\sD_k$ as the conditional
distribution with the protected class being $c_k$. If $\sD$ is the
true underlying distribution, then
\[
\sD_k(x, y) = \sD(x, y \mid c(x, y) = c_k).
\]
Let $\Lambda = \set{\delta_k \colon k \in [p]}$ be the collection of
Dirac measures over the indices $k$ in $[p]$. With this definition, we
define a \emph{good-intent fairness} algorithm as one seeking to
minimize the agnostic loss $\sL_{\sD_\Lambda}$. Thus, the objective of
the algorithm is to minimize the maximum loss incurred on any of the
underlying protective classes and hence does not overfit the data to
any particular model at the cost of others. Furthermore, it does not
degrade the performance of the other classes so long as it does not
affect the loss of the most-sensitive protected category. We further
note that our approach does not reduce bias in the training data
and is useful only for mitigating the training procedure bias.

\section{Learning bounds}
\label{sec:theory}

\arxiv{In this section, we}\icml{We now} present learning guarantees for agnostic
federated learning.
Let $\sG$ denote the family of the losses associated to a hypothesis
set $\sH$:
$\sG = \set{(x, y) \mapsto \ell(h(x), y) \colon h \in \sH}$.  Our
learning bounds are based on the following notion of \emph{weighted
  Rademacher complexity} which is defined for any hypothesis set
$\sH$, vector of sample sizes $\bm = (m_1, \ldots, m_p)$ and mixture
weight $\lambda \in \Delta_p$, by the following expression:
\begin{equation}
\label{eq:WeightedRademacher}
\R_{\bm}(\sG, \lambda) 
= \E_{\substack{S_k \sim \sD_k^{m_k} \\ \bsigma}} \left[\sup_{h \in \sH}
  \sum_{k = 1}^p \frac{\lambda_k}{m_k} \sum_{i = 1}^{m_k} \sigma_{k, i} \,
\ell(h(x_{k, i}), y_{k, i}) \right],
\end{equation}
where $S_k = ((x_{k, 1}, y_{k, 1}), \ldots, (x_{k, m_k}, y_{k, m_k}))$
is a sample of size $m_k$ and
$\bsigma = (\sigma_{k, i})_{\substack{k \in [p], i \in [m_k]}}$ a
collection of Rademacher variables, that is uniformly distributed
random variables taking values in $\set{-1, +1}$.  We also defined the
\emph{minimax weighted Rademacher complexity} for a subset
$\Lambda \subseteq \Delta_p$ by
\begin{equation}
\label{eq:radamacher}
\R_\bm(\sG, \Lambda) = \max_{\lambda \in \Lambda} \R_m(\sG, \lambda).
\end{equation}
Let
$\ov \bm = \frac{\bm}{m} = \big(\frac{m_1}{m}, \ldots,
\frac{m_p}{m} \big)$ denote the empirical
distribution over $\Delta_p$ defined by the sample sizes $m_k$, where
$m = \sum_{k = 1}^p m_k$. 
We define the \emph{skewness} of $\Lambda$ with respect to $\ov \bm$
by
\begin{equation}
\label{eq:skewness}
\s(\Lambda \para \ov \bm)
= \max_{\lambda \in \Lambda} \chi^2 ( \lambda \para \ov \bm) + 1,
\end{equation}
where, for any two distributions $p$ and $q$ in $\Delta_p$, 
the chi-squared divergence $\chi^2(p \para q)$ is given by
$\chi^2(p \para q) = \sum^p_{k = 1} \frac{(p_k - q_k)^2}{q_k}$.
We will also denote by $\Lambda_\epsilon$ a minimum $\epsilon$-cover
of $\Lambda$ in $\ell_1$ distance, that is,
\arxiv{\[}\icml{$}
\Lambda_\epsilon = \argmin_{\Lambda' \in C(\Lambda, \epsilon)} |\Lambda|,
\arxiv{\]}\icml{$}
where $C(\Lambda, \epsilon)$ is a set of distributions $\Lambda'$ such
that for every $\lambda \in \Lambda$, there exists $\Lambda'$ such
that $\sum^p_{k=1} |\lambda_k - \lambda'_k| \leq \epsilon$.

Our first learning guarantee is presented in terms of $\R_\bm(\sG,
\Lambda)$, the skewness parameter $\s(\Lambda \para \ov \bm)$
and the $\e$-cover $\Lambda_\e$.

\begin{theorem}\icml{[Appendix~\ref{app:radamacher}]}
\label{thm:radamacher}
Assume that the loss $\ell$ is bounded by $M > 0$. Fix $\e > 0$ and
$\bm = (m_1, \ldots, m_p)$. Then, for any $\delta > 0$, with
probability at least $1 - \delta$ over the draw of samples $S_k \sim
\sD_k^{m_k}$, \arxiv{the following inequality holds }for all $h \in
\sH$ and $\lambda \in \Lambda$\arxiv{:} \icml{, $\sL_{\sD_\lambda}(h)$
  is upper bounded by}
\[
\arxiv{\sL_{\sD_\lambda}(h) \leq  }
\sL_{\ov \sD_{\lambda}}(h) +
2\R_\bm(\sG, \lambda)  +  M \epsilon + M
\sqrt{\frac{\s(\lambda \para \ov \bm)
}{2 m} \log \frac{| \Lambda_\epsilon |}{\delta}},
\]
where $m = \sum_{k = 1}^p m_k$.
\end{theorem}
\arxiv{\begin{proof}
  The proof is an extension of the standard proofs for Rademacher
  complexity generalization bounds
  \citep{KoltchinskiiPanchenko2002,MohriRostamizadehTalwalkar2012}.
  Fix $\lambda \in \Lambda$. For any sample $S = S_1, \ldots, S_p$,
  define $\Psi(S_1, \ldots, S_p)$ by
\[
\Psi(S_1, \ldots, S_p) = \sup_{h \in \sH} 
\left( \sL_{\sD_\lambda}(h) -  \sL_{\ov \sD_\lambda}(h) \right).
\]
Let $S' = (S'_1, \ldots, S'_p)$ be a sample differing from
$S = (S_1, \ldots, S_p)$ only by point $x'_{k, i}$ in $S'_k$ and
$x_{k, i}$ in $S_k$. Then, since the difference of suprema over the
same set is bounded by the supremum of the differences, we can write
\begin{align*}
\Psi(S') - \Psi(S) 
& = \sup_{h \in \sH} \left( \sL_{\sD_\lambda}(h) -  \sL_{\ov \sD'_\lambda}(h) \right)
- 
\sup_{h \in \sH} \left( \sL_{\sD_\lambda}(h) -  \sL_{\ov \sD_\lambda}(h) \right)\\
& \leq \sup_{h \in \sH} 
\left( \sL_{\sD_\lambda}(h) -  \sL_{\ov \sD'_\lambda}(h) \right)
- 
\left( \sL_{\sD_\lambda}(h) -  \sL_{\ov \sD_\lambda}(h) \right)\\
& \leq \sup_{h \in \sH} 
\sL_{\ov \sD_\lambda}(h) -  \sL_{\ov \sD'_\lambda}(h)\\
& = \sup_{h \in \sH} \sum_{k = 1}^p \frac{\lambda_k}{m_k} \sum_{i = 1}^{m_k} 
\ell(h(x'_{k, i}), y'_{k, i}) - \sum_{k = 1}^p \frac{\lambda_k}{m_k} \sum_{i = 1}^{m_k}
\ell(h(x_{k, i}), y_{k, i}) \\
& = \sup_{h \in \sH} \frac{\lambda_k}{m_k} \Big[ \ell(h(x'_{k, i}),
  y'_{k, i}) - \ell(h(x_{k, i}), y_{k, i}) \Big]\\
& \leq \frac{\lambda_k M}{m_k} .
\end{align*}
Thus, by McDiarmid's inequality, for any $\delta > 0$, the following
inequality holds with probability at least $1 - \delta$ for any
$h \in \sH$:
\[
\sL_{\sD_\lambda}(h) \leq \sL_{\ov \sD_\lambda}(h) + \E \left[ \max_{h
    \in \sH} \sL_{\sD_\lambda}(h) - \sL_{\ov \sD_\lambda}(h) \right] +
M \sqrt{\sum^p_{k = 1}\frac{\lambda^2_k}{2m_k} \log \frac{1}{\delta}}.
\]
Therefore, by the union over $\Lambda_\e$, with probability at least $1 - \delta$, for any
$h \in \sH$ and $\lambda \in \Lambda_\e$ the following holds:
\[
\sL_{\sD_\lambda}(h) \leq  \sL_{\ov \sD_\lambda}(h) + \E \left[
  \max_{h \in \sH}  \sL_{\sD_\lambda}(h) -  \sL_{\ov \sD_\lambda}(h) \right] + 
M \sqrt{\sum^p_{k = 1}\frac{\lambda^2_k}{2m_k} \log \frac{|\Lambda_\e|}{\delta}}.
\]
By definition of $\Lambda_\e$, for any $\lambda \in \Lambda$, there
exists $\lambda' \in \Lambda_\e$ such that $\sL_{\sD_\lambda}(h) \leq
\sL_{\sD_\lambda'}(h) + M \e$. In view of that,
with probability at least $1 - \delta$, for any
$h \in \sH$ and $\lambda \in \Lambda$ the following holds:
\[
\sL_{\sD_\lambda}(h) \leq  \sL_{\ov \sD_\lambda}(h) + \E \left[
  \max_{h \in \sH}  \sL_{\sD_\lambda}(h) -  \sL_{\ov \sD_\lambda}(h) \right] + M \e +
M \sqrt{\sum^p_{k = 1}\frac{\lambda^2_k}{2m_k} \log \frac{|\Lambda_\e|}{\delta}}.
\]
The expectation appearing on the right-hand side can be bounded
following standard proofs for Rademacher complexity upper bounds
(see for example \citep{MohriRostamizadehTalwalkar2012}), leading
to 
\[
\E \left[
  \max_{h \in \sH}  \sL_{\sD_\lambda}(h) -  \sL_{\ov \sD_\lambda}(h)
\right] \leq \R_\bm(\sG, \lambda).
\]
The sum $\sum^p_{k = 1}\frac{\lambda^2_k}{m_k}$ can be expressed in
terms of the skewness of $\Lambda$, using the following equalities:
\begin{align*}
m \sum_{k = 1}^p \frac{\lambda^2_k}{m_k}
= \sum_{k = 1}^p \frac{\lambda^2_k}{\frac{m_k}{m}}
= \sum_{k = 1}^p \frac{\lambda^2_k}{\frac{m_k}{m}} + \sum_{k = 1}^p \frac{m_k}{m}
- 2 \sum_{k = 1}^p \lambda_k + 1
= \sum_{k = 1}^p \frac{(\lambda_k - \frac{m_k}{m})^2}{\frac{m_k}{m}} + 1
= \chi^2(\lambda \para \ov \bm) + 1.
\end{align*}
This completes the proof.

\end{proof}}
It can be proven that the skewness parameter appears in a lower bound
on the generalization bound. We will include that result in the final
version of this paper.  The theorem yields immediately upper bounds
for agnostic losses by taking the maximum over $\lambda \in \Lambda$\arxiv{:
for any $\delta > 0$, with probability at least $1 - \delta$, for any
$h \in \sH$,
\begin{align*}
\sL_{\sD_\Lambda}(h) 
& \leq \max_{\lambda \in \Lambda} \left\{
\sL_{\ov \sD_{\lambda}}(h) + 2\R_\bm(\sG, \lambda) + M \epsilon + M
\sqrt{\frac{\s(\lambda \para \ov \bm) }{2 m} \log \frac{|
    \Lambda_\epsilon |}{\delta}} \right\} \\ & \leq \sL_{\ov
  \sD_{\Lambda}}(h) + \max_{\lambda \in \Lambda} \left\{ 2\R_\bm(\sG,
\lambda) + M \epsilon + M \sqrt{\frac{\s(\lambda \para \ov \bm) }{2 m}
  \log \frac{| \Lambda_\epsilon |}{\delta}} \right\}\\ & \leq \sL_{\ov
  \sD_{\Lambda}}(h) + 2\R_\bm(\sG, \Lambda) + M \epsilon + M
\sqrt{\frac{\s(\Lambda \para \ov \bm) }{2 m} \log \frac{|
    \Lambda_\epsilon |}{\delta}}.
\end{align*}
}\icml{ %in terms of $\R_\bm(\sG, \Lambda)$ and $\s(\Lambda \para \ov
%  \bm)$
.}  The following result shows that, for a family of functions
taking values in $\set{-1, +1}$, the Rademacher complexity
$\R_\bm(\sG, \Lambda)$ can be bounded in terms of the VC-dimension and
the skewness of $\Lambda$.

\begin{lemma}\icml{[Appendix~\ref{app:vc}]}
\label{lem:vc}
Let $\ell$ be a loss function taking values in $\set{-1, +1}$ and such
that the family of losses $\sG$ admits VC-dimension $d$. Then, the
following upper bound holds for the weighted Rademacher complexity of
$\sG$:
\[
\R_\bm(\sG, \Lambda) 
\leq \sqrt{2 \s(\Lambda \para \ov \bm) \frac{d}{m} \log
  \bigg[\frac{em}{d} \bigg]}.
\]
\end{lemma}
\arxiv{\begin{proof}
  For any $\lambda \in \Lambda$, define the set of vectors $A_\lambda$
  in $\Rset^{m}$ by
\[
A_\lambda = \set[\bigg]{\bigg[\frac{\lambda_k}{m_k} \ell(h(x_{k, i}), y_{k,
      i}) \bigg]_{(k, i) \in [p] \times [m_k]} \colon \bx \in \sX^{m}, \by
    \in \sY^{m}}. 
\]
For any $\ba \in A_\lambda$, $\| \ba \|_2 = \sqrt{\sum_{k =
      1}^p m_k \frac{\lambda_k^2}{m_k^2}} = \sqrt{\sum_{k =
      1}^p \frac{\lambda_k^2}{m_k}} \leq
  \sqrt{\frac{\s(\Lambda \para \ov \bm)}{m}}$.
Then, by Massart's lemma, for any $\lambda \in \Lambda$, the following
inequalities hold:
\begin{align*}
\R_{\bm}(\sG, \lambda) 
& = \E_{\substack{S_k \sim \sD_k^{m_k} \\ \bsigma}} \left[\sup_{h \in \sH}
  \sum_{k = 1}^p \frac{\lambda_k}{m_k} \sum_{i = 1}^{m_k} \sigma_{k, i} \,
\ell(h(x_{k, i}), y_{k, i}) \right]\\
& \leq \E_{\bsigma} \left[\sup_{\ba
  \in A}
  \sum_{k = 1}^p \sum_{i = 1}^{m_k} \sigma_{k, i}a_{k, i} \right] \\
& \leq \sqrt{\frac{\s(\Lambda \para \ov \bm)}{m}} \,
  \frac{\sqrt{2 \log |A_\lambda|}}{m}\\
& = \frac{\sqrt{2 \s(\Lambda \para \ov \bm) \log |A_\lambda|}}{m}.
\end{align*}
By Sauer's lemma, the following holds for $m \geq d$:
$|A_\lambda| \leq \left(\frac{em}{d} \right)^d$. Plugging in the
right-hand side in the inequality above completes the proof.

\end{proof}}
Both Lemma~\ref{lem:vc} and the generalization bound of
Theorem~\ref{thm:radamacher} can thus be expressed in terms of the
skewness parameter $\s(\Lambda \para \ov \bm)$. Note that \arxiv{modulo the
skewness parameter, the results look very similar to standard
generalization bounds
\citep{MohriRostamizadehTalwalkar2012}. Furthermore,} when $\Lambda$
contains only one distribution and is the average distribution, that
is $\lambda_k = m_k/m$, then the skewness is equal to one and the
results coincide with the standard guarantees in supervised learning.

Theorem~\ref{thm:radamacher} and Lemma~\ref{lem:vc} also provide
guidelines for choosing the domains and $\Lambda$. When $p$ is large
and $\Lambda = \Delta_p$, then, the number of samples per domain could
be small, the skewness parameter $\s(\Lambda \para \ov \bm) = \max_{1
  \leq k \leq p} \frac{1}{m_k}$ would then be large and the
generalization guarantees for the model would become weaker. We
suggest some guidelines for choosing domains in
\arxiv{Section}\icml{Appendix}~\ref{sec:extensions}. We further note
that for a given $p$, if $\Lambda$ contains distributions that are
close to $\ov \bm$, then the model generalizes well.

The corollary above can be straightforwardly extended to cover the
case where the test samples are drawn from some distribution $\sD$,
instead of $\sD_\lambda$. Define $\ell_1(\sD, \sD_\Lambda)$ by
$\ell_1(\sD, \sD_\Lambda) = \min_{\lambda \in \Lambda} \ell_1(\sD,
\sD_\lambda)$. Then, the following result holds.

\begin{corollary}
\label{cor:unknownd}
Assume that the loss function $\ell$ is bounded by $M$. Then, for any
$\epsilon \geq 0$ and $\delta > 0$, with probability at least
$1 - \delta$, the following inequality holds for all $h \in \sH$:
\begin{align*}
\sL_\sD(h) &  \leq \sL_{\ov \sD_{\Lambda}}(h) +
2\R_\bm(\sG, \Lambda)  +  M\ell_1(\sD, \sD_\Lambda) + M \epsilon \icml{\\ &} + M
\sqrt{\frac{\s(\Lambda \para \ov \bm)
}{2 m} \log \frac{| \Lambda_\epsilon |}{\delta}}.
\end{align*}
\end{corollary}
One straightforward choice of the parameter $\e$ is $\e =
\frac{1}{\sqrt{m}}$, but, depending on $|\Lambda_\e|$ and other tperms
of the bound, more favorable choices may be possible. We conclude this
section by adding that alternative learning bounds can be derived for
this problem, as discussed in Appendix~\ref{app:alternative}.

\section{Algorithm}
\label{sec:algorithm}
\arxiv{
In this section, we introduce a learning algorithm for agnostic
federated learning using the guarantees proven in the previous section
and discuss in detail an optimization solution.}

\subsection{Regularization}

The learning guarantees of the previous section suggest minimizing the
asum of the empirical AFL term
$\sL_{\ov \sD_{\Lambda}}(h)$, a term controlling the complexity of
$\sH$ and a term depending on the skewness parameter.  Observe that,
since $\sL_{\ov \sD_\lambda}(h)$ is linear in $\lambda$, the following
equality holds:
\begin{equation}
\label{eq:convex}
\sL_{\ov \sD_{\Lambda}}(h) = \sL_{\ov \sD_{\conv(\Lambda)}}(h),
\end{equation}
where $\conv(\Lambda)$ is the convex hull of $\Lambda$.  Assume that
$\sH$ is a vector space that can be equipped with a norm
$\| \cdot \|$, as with most hypothesis sets used in learning
applications.  Then, given $\Lambda$ and the regularization parameters
$r \geq 0$ and $\gamma \geq 0$, our learning guarantees suggest
minimizing the regularized loss
$\sL_{\ov \sD_{\Lambda_r}}(h) + \gamma \| h \|$, where $\| \cdot \|$
is a suitable norm controlling the complexity of $\sH$ and where
$\Lambda_r$ is defined by
$\Lambda_r = \set{\lambda \in \conv(\Lambda)\colon 1 +
  \chi^2(\lambda \para \ov \bm) \leq r }$.  This can be equivalently
formulated as the following minimization problem:
\begin{equation}
\label{eq:opt}
\min_{h \in\sH} \max_{\lambda \in \conv(\Lambda)} \sL_{\ov
  \sD_{\lambda}}(h) + \gamma \| h \| - \mu \, \chi^2(\lambda \para \ov \bm),
\end{equation}
where $\mu \geq 0$ is a hyperparameter. This defines our 
algorithm for AFL.

Assume that $\ell$ is a convex function of its first argument.  Then,
$\sL_{\ov \sD_{\lambda}}(h)$ is a convex function of $h$.  Since
$\| h \|$ is a convex function of $h$ for any choice of the norm, for
a fixed $\lambda$, the objective
$\sL_{\ov \sD_{\lambda}}(h) + \gamma \| h \| - \mu \,
\chi^2(\lambda \para \ov \bm)$ is a convex function of $h$.  The
maximum over $\lambda$ (taken in any set) of a family of convex
functions is convex. Thus,
$\max_{\lambda \in \conv(\Lambda)} \sL_{\ov \sD_{\lambda}}(h) + \gamma
\| h \| - \mu \, \chi^2(\lambda \para \ov \bm)$ is a convex function
of $h$ and, when the hypothesis set $\sH$ is a convex, \eqref{eq:opt}
is a convex optimization problem. In the next subsection, we present
an efficient optimization solution for this problem, for which we
prove convergence guarantees.

\subsection{Optimization algorithm}
\label{sec:optimization}

When the loss function $\ell$ is convex, the AFL minmax optimization
problem above can be solved using projected gradient descent or other
instances of the generic mirror descent algorithm
\citep{NemirovskiYudin1983}. However, for large datasets, that is $p$
and $m$ large, this can be computationally costly and typically slow
in practice. \citet*{juditsky2011solving} proposed a stochastic
Mirror-Prox algorithm for solving stochastic variational inequalities,
which would be applicable in our context.  We present a simplified
version of their algorithm for the AFL problem that admits a more
straightforward analysis and that is also substantially easier to
implement.

Our optimization problem is over two sets of parameters, the
hypothesis $h \in \sH$ and the mixture weight $\lambda \in
\Lambda$. In what follows, we will denote by
$w \in \sW \subset \Rset^N$ a vector of parameters defining a
predictor $h$ and will rewrite losses and optimization solutions only
in terms of $w$, instead of $h$. We will use the following notation:
\begin{equation}
\label{eq:newnotation}
\L(w, \lambda) = \sum_{k = 1}^p \lambda_k \L_k(w),
\end{equation}
where $\L_k(w)$ stands for $\sL_{\h \sD_k}(h)$, the empirical
loss of hypothesis $h \in \sH$ (corresponding to $w$) on domain $k$:
\arxiv{\[}\icml{$}
\L_k(w) = \frac{1}{m_k} \sum_{i = 1}^{m_k} \ell(h(x_{k, i}), y_{k,
  i}).
\arxiv{\]}\icml{$}
Since the regularization terms do not make the optimization problem
harder, to simplify the discussion, we will consider the unregularized
version of problem~\eqref{eq:opt}. Thus, we will study the following
problem given by the set of variables $w$:
\begin{equation}
\label{eq:opt_w}
\min_{w \in \sW} \max_{\lambda \in \Lambda} \L(w,\lambda).
\end{equation}

\begin{figure}[t]
\centering
\includegraphics[scale=.28]{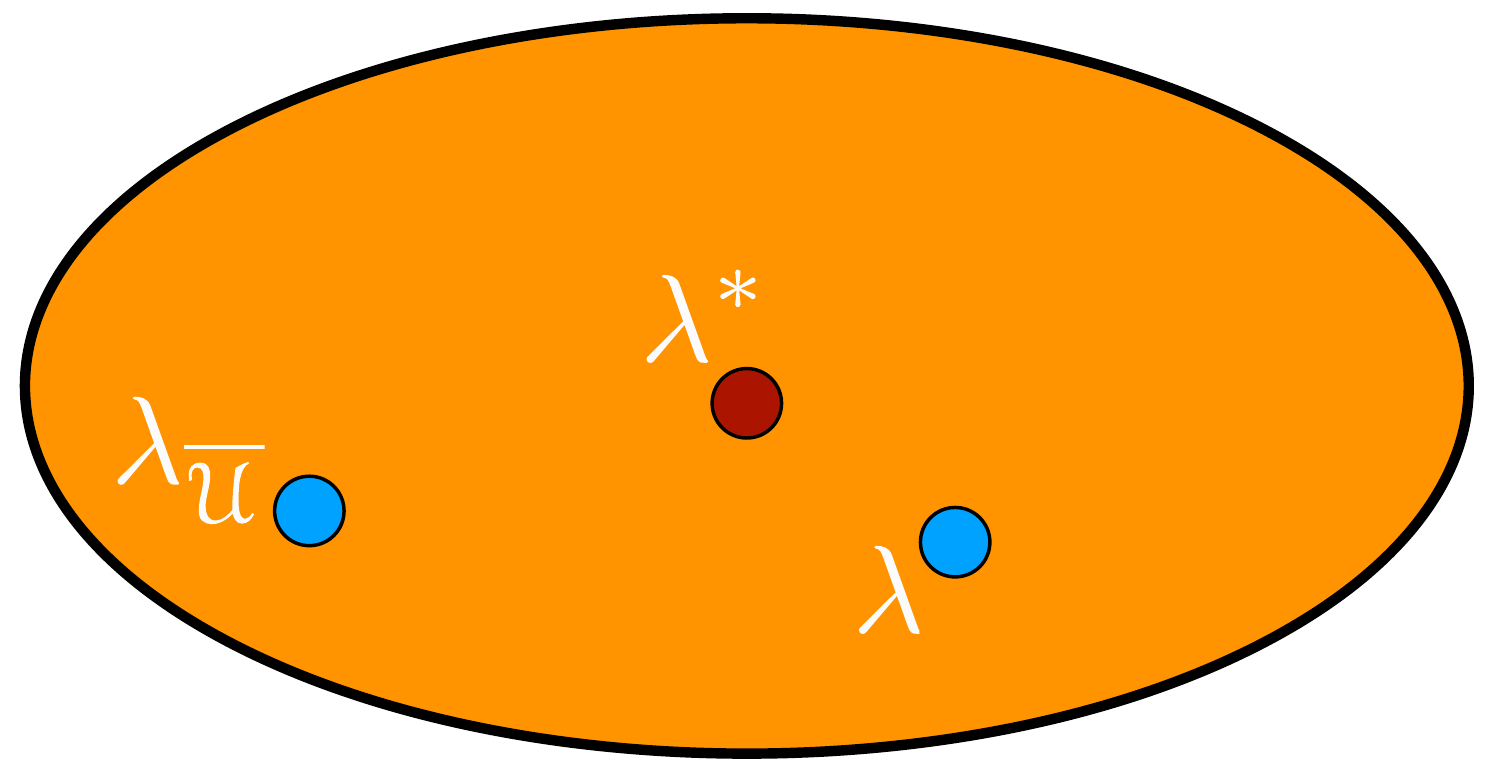}
\caption{Illustration of the positions in $\Lambda$ of $\lambda^*$,
  $\lambda_{\overline \sU}$, the mixture weight corresponding to the
  distribution $\ov \sU$, and an arbitrary $\lambda$. $\lambda^*$
  defines the least risky distribution $\ov \sD_{\lambda^*}$ for
  which to optimize the expected loss.}
\label{fig:lambda_star}
\end{figure}

Observe that problem~\eqref{eq:opt_w} admits a natural game-theoretic
interpretation as a two-player game, where nature selects
$\lambda \in \Lambda$ to maximize the objective, while the learner
seeks $w \in \sW$ minimizing the loss.  We are interested in finding
the equilibrium of this game, which is attained for some $w^*$, the
minimizer of Equation~\ref{eq:opt_w} and $\lambda^* \in \Lambda$, the
hardest domain mixture weights. At the equilibrium, moving $w$ away
from $w^*$ or $\lambda$ from $\lambda^*$, increases the objective
function. Hence, $\lambda^*$ can be viewed as the center of $\Lambda$
in the manifold imposed by the loss function $\L$, whereas
$\overline \sU$, the empirical distribution of samples, may lie
elsewhere, as illustrated by Figure~\ref{fig:lambda_star}.
 
By Equation~\eqref{eq:convex}, using the set $\conv(\Lambda)$ instead
of $\Lambda$ does not affect the solution of the optimization
problem. In view of that, in what follows, we will assume, without
loss of generality, that $\Lambda$ is a convex set. Observe that,
since $\L_k(w)$ is not an average of functions, standard stochastic
gradient descent algorithms cannot be used to minimize this
objective. We will present instead a new stochastic 
gradient-type algorithm for this problem.

\begin{figure}[t]
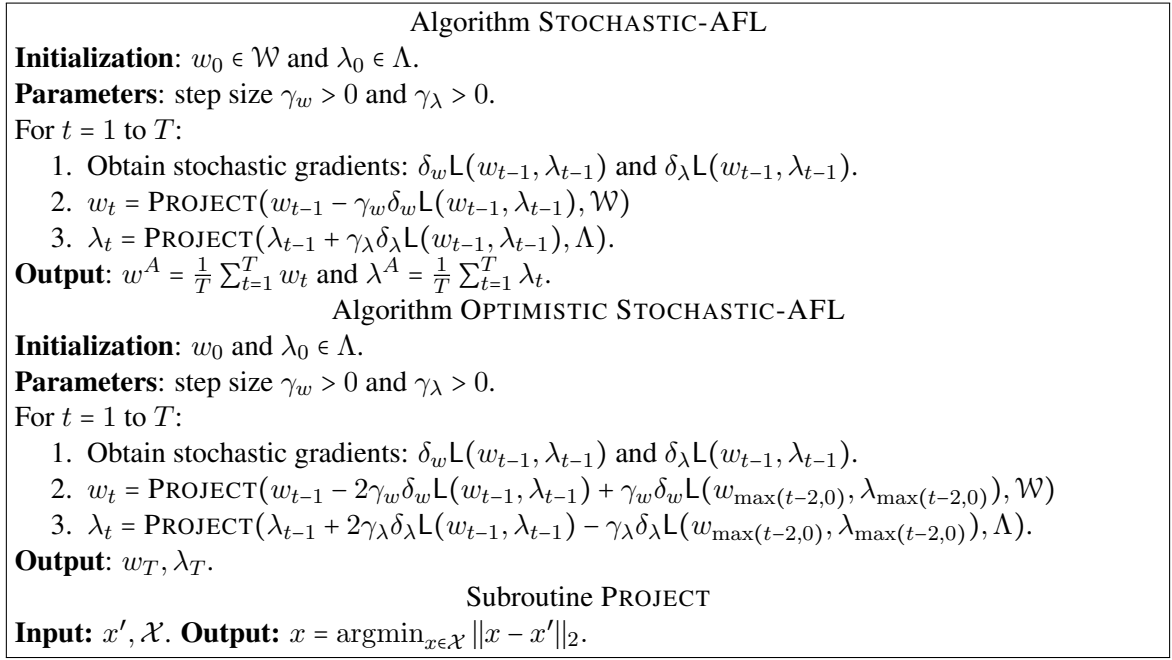

\begin{center}
\fbox{\begin{minipage}{\arxiv{1.0}\icml{0.47}\textwidth}%\icml{{0.5\textwidth}}
\begin{center}
Algorithm \textsc{Stochastic-AFL}
\end{center}
\noindent\textbf{Initialization}: $w_0 \in \sW$ and $\lambda_0 \in \Lambda$. \newline
\noindent\textbf{Parameters}: step size $\gamma_w > 0$ and $\gamma_\lambda > 0$. \newline
\noindent For $t = 1 \ \text{to} \ T$:
\begin{enumerate}
\item Obtain stochastic gradients: $\delta_w
  \L(w_{t-1},\lambda_{t-1})$ and $\delta_\lambda
  \L(w_{t-1},\lambda_{t-1})$.
\item $w_t = \textsc{Project}(w_{t-1} - \gamma_w \delta_w
  \L(w_{t-1},\lambda_{t-1}), \sW)$
\item $\lambda_t = \textsc{Project}(\lambda_{t-1} + \gamma_\lambda
  \delta_\lambda \L(w_{t-1},\lambda_{t-1}), \Lambda)$.
\end{enumerate}
\noindent \textbf{Output}: $w^A = \frac{1}{T} \sum^T_{t=1} w_t $ and $\lambda^A = \frac{1}{T} \sum^T_{t=1} \lambda_t $.
\arxiv{
\begin{center}
Algorithm \textsc{Optimistic Stochastic-AFL}
\end{center}
\noindent\textbf{Initialization}: $w_0$ and $\lambda_0 \in \Lambda$.  \newline
\noindent\textbf{Parameters}: step size $\gamma_w > 0$ and $\gamma_\lambda > 0$. \newline
\noindent For $t = 1 \ \text{to} \ T$:
\begin{enumerate}
\item Obtain stochastic gradients: $\delta_w
  \L(w_{t-1},\lambda_{t-1})$ and $\delta_\lambda
  \L(w_{t-1},\lambda_{t-1})$.
\item $w_t = \textsc{Project}(w_{t-1} - 2\gamma_w \delta_w
  \L(w_{t-1},\lambda_{t-1}) + \gamma_w \delta_w \L(w_{\max(t-2,
    0)},\lambda_{\max(t-2, 0)}), \sW)$
\item $\lambda_t = \textsc{Project}(\lambda_{t-1} + 2\gamma_\lambda
  \delta_\lambda \L(w_{t-1},\lambda_{t-1}) - \gamma_\lambda
  \delta_\lambda \L(w_{\max(t-2, 0)},\lambda_{\max(t-2, 0)}),
  \Lambda)$.
\end{enumerate}
\noindent\textbf{Output}: $w_T, \lambda_T$.
}
\begin{center}
Subroutine \textsc{Project}
\end{center}
\noindent\textbf{Input:} $x',\cX$.
\noindent\textbf{Output:} $x = \argmin_{x \in \cX} ||x - x'||_2.$
\end{minipage}}
\end{center}
\caption{Pseudocodes of the \textsc{Stochastic-AFL} and
\textsc{Optimistic Stochastic-AFL} algorithms.}
\label{fig:algo}
\end{figure}

Let $\nabla_w \L(w, \lambda)$ denote the gradient of the
loss function with respect to $w$ and $\nabla_\lambda \L(w, \lambda)$
the gradient with respect to $\lambda$. Let $\delta_w \L(w,
\lambda)$, and $\delta_{\lambda} \L(w, \lambda)$ be unbiased estimates
of the gradient, that is,
\[
\E_\delta[\delta_{\lambda} \L(w, \lambda)] = \nabla_\lambda \L(w,
\lambda) \text{\, \arxiv{and}\icml{,} \,}\E_\delta[\delta_w \L(w, \lambda)] = \nabla_w
\L(w, \lambda).
\]
We first give an optimization algorithm \textsc{Stochastic-AFL} for
the AFL problem, assuming access to such unbiased estimates.  The
pseudocode of the algorithm is given in Figure~\ref{fig:algo}.  At
each step, the algorithm computes a stochastic gradient with respect
to $\lambda$ and $w$ and updates the model accordingly. It then
projects $\lambda$ to $\Lambda$ by computing a value in $\Lambda$ via
convex minimization. If $\Lambda$ is the full simplex, then there is a
near-linear time algorithm for this projection
\cite{wang2013projection}. It then repeats the process for $T$ steps
and return the average of the weights. We provide guarantees for this
algorithm in terms of the variance of the stochastic gradients when
the loss function $\L$ is convex and when the set of $w$s, $\sW$, is
a compact set.

In the above analysis and in algorithm description in \ref{fig:algo},
we have ignored the regularization term. If the objective contains a
regularization term such as Equation~\ref{eq:opt}, then for
$\lambda_k$, the regularization term yields a derivative of $- 2 \gamma
\lambda_k / \ov \bm_k$, which can be added to $\delta_{\lambda} \L(w,
\lambda)$ in Step $3$ in Algorithm~\ref{fig:algo}.

There are several natural candidates for the sampling method defining
stochastic gradients. We highlight two techniques: \textsc{PerDomain
  gradient} and \textsc{Weighted gradient}. We analyze the time
complexity and give bounds on the variance for both techniques in
Lemmas~\ref{lem:perdomain} and \ref{lem:weighted} respectively.

Recently, \cite{rakhlin2013optimization} and
\cite{daskalakis2017training} gave an optimistic gradient descent
algorithm for minimax optimizations. Our algorithm can also be
modified to derive a stochastic optimistic algorithm, which we
\arxiv{refer to as \textsc{Optimistic-Stochastic-AFL}.  The pseudocode
  of this algorithm is also given in Figure~\ref{fig:algo}. However,
  the convergence analysis we present in the next section does not
  cover this algorithm.} \icml{provide in the full version of the
  paper.}

\subsection{Analysis}

Throughout this section, for simplicity, we adopt the notation
introduced for Equation~\ref{eq:newnotation}. Our convergence
guarantees hold under the following assumptions, which are similar to
those adopted for the convergence proof of gradient descent-type
algorithms.

\begin{properties}
\label{prop:1}
Assume that the following properties hold for the loss function $\L$
and sets $\sW$ and $\Lambda \subseteq \Delta_p$:
\begin{enumerate}

\item \emph{Convexity}: $w  \mapsto \L(w, \lambda)$ is convex for any
  $\lambda \in \Lambda$.

\item \emph{Compactness}:
  $\max_{\lambda \in \Lambda} \| \lambda \|_2 \leq R_\Lambda$ and
  $\max_{w \in \sW} \| w \|_2 \leq R_{\sW}$, for some
  $R_\Lambda > 0$ and $R_{\sW} > 0$.

\item \emph{Bounded gradients}:
  $\| \nabla_w \L(w, \lambda) \|_2 \leq G_w$ and
  $\| \nabla_\lambda \L(w, \lambda) \|_2 \leq G_\lambda$ for all
  $w \in \sW$ and $\lambda \in \Lambda$.

\item \emph{Stochastic variance}:
  $\E[\| \delta_w \L(w, \lambda) - \nabla_w \L(w, \lambda) \|_2^2]
  \leq \sigma^2_w$ and
  $\E[\| \delta_\lambda \L(w, \lambda) - \nabla_\lambda \L(w,
  \lambda) \|_2^2] \leq \sigma^2_\lambda$ for all $w \in \sW$ and
  $\lambda \in \lambda$.

\item \emph{Time complexity}: $U_w$ denotes the time complexity of
  computing $\delta_w \L(w, \lambda)$, $U_\lambda$ that of computing
  $\delta_\lambda \L(w, \lambda)$, $U_p$ that of the projection, and
  $d$ denotes the dimensionality of $\sW$.

\end{enumerate}
\end{properties}
\begin{theorem}\icml{[Appendix~\ref{app:convergence}]}
\label{thm:convergence}
Assume that the Properties~\ref{prop:1} hold. Then, for the steps
sizes $\gamma_w = \frac{2R_{\sW}}{\sqrt{T(\sigma^2_w + G^2_w)}}$ and
$\gamma_\lambda = \frac{2R_\Lambda}{\sqrt{T(\sigma^2_\lambda +
    G^2_\lambda)}}$, the following guarantee holds for
\textsc{Stochastic-AFL}:
\[
\E \left[\max_{\lambda \in \Lambda} \L(w^A, \lambda)
- \min_{w \in \sW} \max_{\lambda \in \Lambda}  \L(w, \lambda)  \right] \arxiv{\leq}
\icml{\] is upper bounded by \[} 
\frac{3R_{\sW} \sqrt{(\sigma^2_w + G^2_w)}}{\sqrt{T}} + \frac{3R_\Lambda \sqrt{(\sigma^2_\lambda + G^2_\lambda)}}{\sqrt{T}}.
\]
and the time complexity of the algorithm is in
$\mathcal{O}((U_\lambda + U_w + U_p + d + k) T)$.
\end{theorem}
\arxiv{\begin{proof}
  The time complexity of the algorithm follows the definitions of the
  complexity terms $U_\lambda$, $U_w$, and $U_p$ the dimension $d$ in
  Properties~\ref{prop:1}. To prove the convergence guarantee, we make
  a series of reductions. Let $w^A$ and $\lambda^A$ be a solution
  returned by the algorithm. First observe that since $\L$ is convex
  in $w$ and linear and thus concave in $\lambda$, by the generalized
  von Neumann's theorem, the following holds:
\begin{align*}
\max_{\lambda \in \Lambda} \L(w^A, \lambda)
- \min_{w \in \sW} \max_{\lambda \in \Lambda}  \L(w, \lambda) 
& = 
\max_{\lambda \in \Lambda} \L(w^A, \lambda)
- \max_{\lambda \in \Lambda} \min_{w \in \sW} \L(w, \lambda)  & \text{(von Neumann's minimax)}\\
& \leq \max_{\lambda \in \Lambda} \set[\Big]{ \L(w^A, \lambda)
- \min_{w \in \sW} \L(w, \lambda^A)} & \text{(subadd. of $\max$)}\\
& = \max_{\substack{\lambda \in \Lambda\\ w \in \sW}}
  \set[\Big]{\L(w^A, \lambda)
- \L(w, \lambda^A) } \\
& \leq \frac{1}{T} \max_{\substack{\lambda \in \Lambda\\ w \in \sW}}
  \set[\bigg]{ \sum^T_{t = 1} \L(w_t,\lambda) - \L(w, \lambda_t)
  }.  & \text{(convexity in $w$ and lin. in $\lambda$)}
\end{align*}
Next, since the function is linear in $\lambda$ and convex in $w$,
\begin{align*}
 \L(w_t, \lambda)
- \L(w, \lambda_t)  
& = \L(w_t,\lambda) - \L(w_t, \lambda_t) + \L(w_t, \lambda_t)
- \L(w, \lambda_t) \\
& \leq (\lambda - \lambda_t) \nabla_\lambda \L(w_t, \lambda_t)
 + (w_t - w) \nabla_w \L(w_t, \lambda_t) \\
 & \leq (\lambda - \lambda_t) \delta_\lambda \L(w_t, \lambda_t)
  + (w_t - w) \delta_w \L(w_t, \lambda_t) \\& + 
  (\lambda - \lambda_t) (\nabla_\lambda \L(w_t, \lambda_t) - \delta_\lambda \L(w_t, \lambda_t)) 
  + (w_t - w) (\nabla_w  \L(w_t, \lambda_t)- \delta_w  \L(w_t, \lambda_t)). 
\end{align*}
In view of these inequalities, by the subadditivity of $\max$, the
following inequality holds:
\begin{align*}
&  \max_{\substack{\lambda \in \Lambda\\ w \in \sW}}
  \set[\Big]{ \sum^T_{t = 1} \L(w_t,\lambda) - \L(w, \lambda_t)
  } \\
& \leq \max_{\substack{\lambda \in \Lambda\\ w \in \sW}} \sum^T_{t=1}(\lambda - \lambda_t) \delta_\lambda \L(w_t, \lambda_t)
  + (w_t - w) \delta_w \L(w_t, \lambda_t) \\
&  + \max_{\substack{\lambda \in \Lambda\\ w \in \sW}} \sum^T_{t=1} \lambda (\nabla_\lambda \L(w_t, \lambda_t) - \delta_\lambda \L(w_t, \lambda_t)) 
   - w (\nabla_w  \L(w_t, \lambda_t)- \delta_w  \L(w_t, \lambda_t)) \\
   & + \sum^T_{t=1} \lambda_t (\nabla_\lambda \L(w_t, \lambda_t) - \delta_\lambda \L(w_t, \lambda_t)) 
   - w_t (\nabla_w  \L(w_t, \lambda_t)- \delta_w  \L(w_t, \lambda_t)).
\end{align*}
We now bound each of the terms above separately. 
For the first term, observe that for any $w \in \sW$, 
\begin{align*}
&  \sum^T_{t=1}(w_t - w) \delta_w \L(w_t, \lambda_t) \\
 & =  \frac{1}{2\gamma_w} \sum^T_{t=1} \|(w_t - w)\|^2_2 + \gamma^2_w\| \delta_w \L(w_t, \lambda_t)\|^2_2 -  \|(w_t - \gamma_w \delta_w \L(w_t, \lambda_t) - w)\|^2_2 \\
 & \leq \frac{1}{2\gamma_w} \sum^T_{t=1} \|(w_t - w)\|^2_2 +
   \gamma^2_w\| \delta_w \L(w_t, \lambda_t)\|^2_2 -  \|(w_{t+1} -
   w)\|^2_2 & \text{(property of projection)}\\
 & = \frac{1}{2\gamma_w} \|(w_1 - w)\|^2_2 - \|(w_{T + 1} - w)\|^2_2 +
   \frac{\gamma_w}{2} \sum^T_{t=1} \| \delta_w \L(w_t,
   \lambda_t)\|^2_2 & \text{(telescoping sum)}\\
 & \leq \frac{1}{2\gamma_w} \|(w_1 - w)\|^2_2 + \frac{\gamma_w}{2} \sum^T_{t=1} \| \delta_w \L(w_t, \lambda_t)\|^2_2 \\
 & \leq \frac{2R^2_{\sW}}{\gamma_w} + \frac{\gamma_w}{2} \sum^T_{t=1}
   \| \delta_w \L(w_t, \lambda_t)\|^2_2\\
 & \leq \frac{2R^2_{\sW}}{\gamma_w} + \frac{\gamma_w}{2} \sum^T_{t=1}
   \| \delta_w \L(w_t, \lambda_t) - \nabla_w \L(w_t, \lambda_t) +
   \nabla_w \L(w_t, \lambda_t) \|^2_2.
 \end{align*}
Since the right-hand side does not depend on $w$, taking the maximum
of both sides over $w \in \sW$ and the expectation yields
\begin{align*}
 \E\left[\max_{w \in \sW} \sum^T_{t=1}(w_t - w) \delta_w \L(w_t, \lambda_t)
   \right] \leq \frac{2R^2_{\sW}}{\gamma_w} +  \frac{\gamma_w
   T\sigma^2_w}{2} +  \frac{T \gamma_w G^2_w}{2},
\end{align*}
using the following identity:
\begin{align*}
& \E \left[ \| \delta_w \L(w_t, \lambda_t) - \nabla_w \L(w_t, \lambda_t) +
   \nabla_w \L(w_t, \lambda_t) \|^2_2 \right]\\
& = \E \left[ \| \delta_w \L(w_t, \lambda_t) - \nabla_w \L(w_t,
  \lambda_t) \|^2 \right] -2 \E \left[ \delta_w \L(w_t, \lambda_t) -
  \nabla_w \L(w_t, \lambda_t) \right] \cdot \nabla_w \L(w_t,
  \lambda_t) + \| \nabla_w \L(w_t, \lambda_t) \|^2_2\\
& = \E \left[ \| \delta_w \L(w_t, \lambda_t) - \nabla_w \L(w_t,
  \lambda_t) \|^2 \right] + \| \nabla_w \L(w_t, \lambda_t) \|^2_2.
\end{align*}
Similarly, using the projection property, the following inequality can
be shown:
\[
 \E\left[\max_{\lambda \in \Lambda} \sum^T_{t=1}(\lambda - \lambda_t) \delta_\lambda \L(w_t, \lambda_t) \right]
 \leq \frac{2R^2_{\Lambda}}{\gamma_\lambda} + \frac{\gamma_\lambda T\sigma^2_\lambda}{2} + \frac{T \gamma_\lambda G^2_\lambda}{2}.
\]
For the second term, by the Cauchy-Schwarz inequality, we can write
\begin{align*}
\max_{\lambda \in \Lambda} \sum^T_{t=1} \lambda (\nabla_\lambda
  \L(w_t, \lambda_t) - \delta_\lambda \L(w_t, \lambda_t)) 
& \leq R_\Lambda \| \sum^T_{t=1}\nabla_\lambda \L(w_t, \lambda_t) -
  \delta_\lambda \L(w_t, \lambda_t) \|_2 \\
& \leq R_\Lambda \sum^T_{t=1} \| \nabla_\lambda \L(w_t, \lambda_t) -
  \delta_\lambda \L(w_t, \lambda_t) \|_2.
\end{align*}
Taking the expectation of both sides and using Jensen's inequality yields
\[
\E\left[  \max_{\lambda \in \Lambda} \sum^T_{t=1} \lambda (\nabla_\lambda \L(w_t, \lambda_t) - \delta_\lambda \L(w_t, \lambda_t))\right]
\leq R_\Lambda \sqrt{T} \sigma_\lambda.
\]
Similarly, we obtain the following:
\[
\E\left[  \max_{w \in \sW} w \nabla_w  \L(w_t, \lambda_t)- \delta_w  \L(w_t, \lambda_t) \right] \leq 
R_{\sW} \sqrt{T} \sigma_w.
\]
For the third term, observe that the stochastic gradients at time $t$
are unbiased, conditioned on $\lambda_t$, and $w_t$, hence,
\[
\E\left[\sum^T_{t=1} \lambda_t (\nabla_\lambda \L(w_t, \lambda_t) - \delta_\lambda \L(w_t, \lambda_t)) 
   - w_t (\nabla_w  \L(w_t, \lambda_t)- \delta_w  \L(w_t, \lambda_t))\right] =0.
\]
Combining the upper bounds just derived gives:
\begin{multline*}
\E\left[\max_{\lambda \in \Lambda} \L(w^A, \lambda)
- \min_{w \in \sW} \max_{\lambda \in \Lambda}  \L(w, \lambda)\right] \\
\leq  \frac{2R^2_{\sW}}{T\gamma_w} + \frac{\gamma_w (\sigma^2_w +
  G^2_w)}{2} + \frac{2R^2_{\Lambda}}{T\gamma_\lambda} +
  \frac{\gamma_\lambda (\sigma^2_\lambda + G^2_\lambda)}{2}  +
  \frac{R_{\sW} \sigma_w }{\sqrt{T}} + \frac{R_\Lambda\sigma_\lambda}{ \sqrt{T} }.
\end{multline*}
Setting $\gamma_w = \frac{2 R_{\sW}}{\sqrt{T( (\sigma^2_w + G^2_w))}}$
and $\gamma_\lambda = \frac{2 R_\Lambda}{\sqrt{T( (\sigma^2_\lambda +
    G^2_\lambda))}}$ to minimize this upper bound
completes the proof.

\end{proof}}

\subsection{Stochastic gradients}

The convergence results of Theorem~\ref{thm:convergence} depend on the
variance of the stochastic gradients.  \arxiv{Thus, before proceeding to the
results, we first compute the gradients with respect to $w$ and
$\lambda$. Let $\L_{k, i}(w) = \ell(h(x_{k, i}, y_{k, i}))$. 
For any $w \in \sW$, $\lambda \in \Lambda$ and $k \in [p]$, 
the gradient with respect to $w$ is given by
\[
\nabla_w \L(w, \lambda) = \sum_{k = 1}^p \frac{\lambda_k}{m_k}
\sum^{m_k}_{i = 1} \nabla_w \L_{k, i}(w).
\]
For any $w \in \sW$, $\lambda \in \Lambda$ and $k \in [p]$, the 
gradient with respect to $\lambda_k$ is given by
\[
[\nabla_\lambda \L(w, \lambda)]_k 
= \frac{1}{m_k} \sum^{m_k}_{i = 1} \L_{k, i}(w)
= \L_k(w).
\]}
We first discuss the stochastic gradients for $\lambda$. Notice that
the gradient for $\lambda$ is independent of $\lambda$. Thus, a
natural choice for the stochastic gradient with respect to $\lambda$ is
based on uniformly sampling a domain $K \in [p]$ and then sampling $x_{K, i}$
from domain $K$. This leads to the definition of the stochastic
gradient $\delta_\lambda \L(w, \lambda)$ shown in Figure~\ref{fig:SGD}.
The following lemma bounds the variance for that
definition of $\delta_\lambda \L(w, \lambda)$.
\begin{lemma}\icml{[Appendix~\ref{app:stochastic_lambda}]}
\label{lem:stochastic_lambda}
  The stochastic gradient $\delta_\lambda \L(w, \lambda)$ is
  unbiased. Further, if the loss function is bounded by $M$, then the
  following upper bound holds for the variance of
  $\delta_\lambda \L(w, \lambda)$:
\[
\sigma^2_\lambda 
= \max_{w \in \sW, \lambda \in \Lambda} \var(\delta_\lambda \L(w, \lambda)) \leq p^2 M^2.
\]
\end{lemma}
\arxiv{
\begin{proof}
  The unbiasedness of $\delta_\lambda \L(w, \lambda)$ follows directly
  its definition. For the variance, observe that, for index
  $k \in [p]$, since the probability of not drawing domain $k$ is $(1
  - \frac{1}{p})$, the variance is given by the following
\begin{align*}
\var_k[\delta_\lambda \L(w, \lambda)]
& = \bigg[ 1 - \frac{1}{p} \bigg] [0 - \L_k(w)]^2 + \frac{1}{p} \sum_{k = 1}^p \frac{1}{m_k} \sum_{i = 1}^{m_k} \left[ p \L_{k,
  i}(w) - \L_k(w)  \right]^2 \\
& \leq \bigg[ 1 - \frac{1}{p} \bigg] M^2 + \frac{1}{p} \sum_{k = 1}^p
  \frac{1}{m_k} \sum_{i = 1}^{m_k} [pM]^2 = pM^2.
\end{align*}
Summing over all indices from $k \in [p]$ completes the proof.

\end{proof}}
If the above variance is too high, then we can sample one $J_k$ for
every domain $k$. This is same as computing the gradient of a batch
and reduces the variance by a factor of $p$.

\begin{figure}[t]
\begin{center}
\fbox{\begin{minipage}{\arxiv{1.0}\icml{0.47}\textwidth}
\begin{center}
Stochastic gradient for $\lambda$.
\end{center}
\begin{enumerate}
\item Sample $K \sim [p]$, according to the uniform distribution.\\ 
Sample $I_K \sim [m_K]$, according to the uniform distribution. 
\item $\delta_\lambda \L(w, \lambda)$ such that $[\delta_\lambda
  \L(w, \lambda)]_K = p \L_{K, I_K}(w)$ and for all $k \neq K$, $[\delta_\lambda \L(w, \lambda)]_k = 0$.
\end{enumerate}
\noindent\textbf{Output}: $\delta_w \L(w, \lambda), \delta_\lambda(w, \lambda)$. 
\begin{center}
\textsc{PerDomain}-stochastic gradient for $w$.
\end{center}
\begin{enumerate}
\item For $k \in [p]$, sample $J_k \sim [m_k]$, according to the uniform distribution.
\item $\delta_w \L(w, \lambda) = \sum_{k = 1}^{p}  \lambda_k
  \nabla_w L_{k, J_k}(w, h)$.
\end{enumerate}
\begin{center}
\textsc{Weighted}-stochastic gradient for $w$
\end{center}
\begin{enumerate}
\item Sample $K \sim [p]$ according to the distribution $\lambda$.\\
Sample $J_K \sim [m_k]$, according to the uniform distribution.
\item $\delta_w \L(w, \lambda) = \nabla_w \L_{K, J_K}(w)$.
\end{enumerate}
\end{minipage}}
\end{center}
\caption{Definition of the stochastic gradients with respect to
  $\lambda$ and $w$.}
\label{fig:SGD}
\end{figure}

The gradient with respect to $w$ depends both on $\lambda$ and
$w$. There are two natural stochastic gradients: 
the \textsc{PerDomain}-stochastic gradient and 
the \textsc{Weighted}-stochastic gradient. For a 
PerDomain-stochastic gradient, we sample an element uniformly from
$[m_k]$ for each $k \in [p]$. For the \textsc{Weighted}-stochastic
gradient, we sample a domain according to $\lambda$ and sample an
element out of it. \arxiv{To bound the variance of these two stochastic
gradients, we need a few definitions.

\begin{definition}
The following definitions are used:
\begin{itemize}

\item the \emph{intra-domain variance} 
  with respect to $w$ is defined as follows:
\[
\sigma^2_I(w)= \max_{w \in \sW, k \in [p]}
\frac{1}{m_k}\sum^{m_k}_{j = 1} \left[\nabla_w L_{k,j}(w) - \nabla_w L_{k}(w)  \right]^2.
\]

\item the \emph{outer-domain variance}
  with respect to $w$ is defined as follows:
\[
\sigma^2_O(w) = \max_{w\in \sW, \lambda \in \Lambda} \sum^p_{k = 1}
\lambda_k 
\left[\nabla_w \L_k(w) - \nabla_w \L(w, \lambda)  \right]^2.
\]

\item the \emph{time complexity} of computing the loss and gradient
  with respect to $w$ for a single sample is denoted by $U$.

\end{itemize}
\end{definition}

With these definitions, we can }\icml{We can now }bound the variance of both
\textsc{PerDomain} and \textsc{Weighted} stochastic gradients.
\begin{lemma}
\label{lem:perdomain}\icml{[Appendix~\ref{app:perdomain}]}
\textsc{PerDomain} stochastic gradient is unbiased and runs in time $p
U + \mathcal{O}(p \log m)$ and the variance satisfy,
\arxiv{\[}\icml{$}
\sigma^2_w \leq R_{\Lambda} \sigma^2_I(w)\arxiv{.}\icml{,}
\arxiv{\]}\icml{$}
\icml{where \[\sigma^2_I(w)= \max_{w \in \sW, k \in [p]}
  \frac{1}{m_k}\sum^{m_k}_{j = 1} \left[\nabla_w L_{k,j}(w) - \nabla_w
    L_{k}(w) \right]^2.\]}
\end{lemma}
\arxiv{\begin{proof}
  The time complexity and the unbiasedness follow from the
  definitions. We now bound the variance.  Since $\nabla_w \L_{k, J_k}$
  is an unbiased estimate of $\nabla_w \L_k(w)$ and we have:
\begin{align*}
\var[\delta_w] = \sum_{k = 1}^p \lambda^2_k  \var\left[
\nabla_w \L_{k, J_k}(w) - \nabla_w \L_k(w)  \right] 
\leq \sum_{k = 1}^p \lambda^2_k  \sigma^2(w, I) 
\leq R_{\Lambda} \sigma^2_I(w).
\end{align*}
This completes the proof.

\end{proof}}
\begin{lemma}\icml{[Appendix~\ref{app:weighted}]}
\label{lem:weighted}
\textsc{Weighted} stochastic gradient is unbiased and runs in time $U
+ \mathcal{O}(k + \log n )$ and the variance satisfy the following
inequality: \arxiv{\[}\icml{$} \sigma^2_w \leq \sigma^2_I(w) +
\sigma^2_O(w)\arxiv{.}\icml{,} \arxiv{\]}\icml{$}
\icml{where \[\sigma^2_O(w) = \max_{w\in \sW, \lambda \in \Lambda}
  \sum^p_{k = 1} \lambda_k \left[\nabla_w \L_k(w) - \nabla_w \L(w,
    \lambda) \right]^2\] and $\sigma^2_I(w)$ is defined in
  Lemma~\ref{lem:perdomain}.}
\end{lemma}
\arxiv{
\begin{proof}  The time complexity and the unbiasedness follow from the
  definitions. We now bound the variance. By definition for any
  $w, \lambda$,
\begin{align*}
\var(\delta_w) &
 = \sum^p_{k = 1} \frac{\lambda_k}{m_k} \sum^{m_k}_{j = 1} 
 \left( \nabla_w \L_{k, j}(w) - \L(w, \lambda) \right)^2 \\
& = \sum^p_{k = 1} \frac{\lambda_k}{m_k} \sum^{m_k}_{j = 1} 
 \left( \nabla_w \L_{k, j}(w) - \L_k(w) \right)^2 + 
 \sum^p_{k = 1} \lambda_k (\L_k(w) - \L(w, h))^2 \\
& \leq \sigma^2_I(w) + \sigma^2_O(w),
\end{align*}
where the second equality follows from the unbiasedness of the stochastic gradients.

\end{proof}
}
Since $R_{\Lambda}\leq 1$, at first glance, the above two lemmas may
suggest that \textsc{PerDomain} stochastic is always better than
\textsc{Weighted} stochastic gradient. Note, however, that the time
complexities of the algorithms is dominated by $U$ and thus, the time
complexity of \textsc{PerDomain}-stochastic gradient is roughly $k$
times larger than that of \textsc{Weighted}-stochastic
gradient. Hence, if $k$ is small, it is preferable to choose the
\textsc{PerDomain}-stochastic gradient.

For large values of $p$, to do a fair comparison, we need to average
$p$ independent copies of the \textsc{Weighted}-stochastic gradient,
which we refer to as $p$-\textsc{Weighted}, and compare it with the
\textsc{PerDomain}-stochastic gradient. Since the variance of average of
$p$ i.i.d.\ random variables is $1/p$ times the individual variance,
by Lemma~\ref{lem:weighted}, \arxiv{the following holds:
\[}\icml{$}
\var(\textsc{$k$-Weighted}) = \frac{\sigma^2_I(w) + \sigma^2_O(w)}{p}.
\arxiv{\]}\icml{$}
Further, observe that $R_{\Lambda} = \max_{\lambda \in \Lambda}
\sum^{p}_{k = 1} \lambda^2_k \geq \frac{1}{p}$. Thus,
\arxiv{\[}\icml{$}
\var(\textsc{PerDomain}) \geq \frac{\sigma^2_I(w)}{p}.
\arxiv{\]}\icml{$}
Hence, the right choice of the stochastic variance of $w$ depends on
the application. If all domains are roughly equally weighted, then we
have $R(\Lambda) \approx \frac{1}{p}$ and the \textsc{PerDomain}-variance is a
more favorable choice. Otherwise, if $\sigma^2_O(w)$ is small, then
the \textsc{Weighted}-stochastic gradient is more favorable.

\section{Experiments}
\label{sec:experiments}
\begin{table*}[t]
  \centering
  \caption{Test accuracy of the train model on various domains, as a
      function of training loss for the adult dataset.\arxiv{ Of all the model, domain agnostic
      model that minimizes $\sL_{\sD_{\Lambda}}$ has the best accuracy
      on the worst domain. All experiments are averaged over $50$
      runs.}}
\vskip .15in
  \begin{tabular}{l c c c c} % \hline
    Training loss &  $\sU$  & \texttt{doctorate} & \texttt{non-doctorate} & $\sD_{\Lambda}$ \\ \hline
    $\sL_{\texttt{doctorate}}$ & $53.35 \pm 0.91$  & $73.58 \pm 0.48$ & $53.12 \pm 0.89$ & $53.12 \pm 0.89$ \\
   $\sL_{\texttt{non-doctorate}}$ & $82.15 \pm 0.09$  & $69.46 \pm 0.29$ & $82.29 \pm 0.09$  & $69.46 \pm 0.29$  \\
    $\sL_{\h \sU}$ &  $82.10 \pm 0.09$  & $69.61\pm 0.35$ & $82.24 \pm 0.09$ & $69.61\pm 0.35$ \\
    $\sL_{\sD_{\Lambda}}$ &  $80.10 \pm 0.39$ & $71.53 \pm 0.88$ & $80.20 \pm 0.40$& $71.53 \pm 0.88$   \\ %\hline
  \end{tabular}
\label{tab:adult}
\end{table*}

\begin{table*}[t]
  \centering
  \caption{Test accuracy of the train model on the different
      clothing classes, as a function of training loss for the Fashion MNIST dataset.\arxiv{ Of the two
      models, the domain agnostic model that minimizes
      $\sL_{\sD_{\Lambda}}$ has the best accuracy overall and on the
      worst domain. All experiments are averaged over $50$ runs.}}
\vskip .15in
  \begin{tabular}{l c c c c c c } % \hline
%    \multicolumn{2}{|c|}
{Training loss} &  $\sU$  & \texttt{shirt} & \texttt{pullover} & \texttt{T-shirt/top} & $\sD_{\Lambda}$ \\ \hline
%    \multirow{2}{*}
{$\sL_{\h \sU}$} 
%& accuracy 
& $81.8 \pm 1.3$ & $71.2 \pm 7.8$ & $87.8 \pm 6.0$ & $86.2 \pm 4.9$ &
$71.2 \pm 7.8$ \\
%    & loss &  $0.440 \pm 0.025$  &  $0.643 \pm 0.169$  &  $0.335 \pm 0.127$  &  $0.344 \pm 0.088$  &  $0.643 \pm 0.169$ \\ 
%\hline
%    \multirow{2}{*}
{$\sL_{\sD_{\Lambda}}$} & 
%accuracy &  
$82.3 \pm 0.9$  &  $74.5 \pm 6.0$  &  $87.6 \pm 4.5$  &  $84.9 \pm 4.4$  &  $74.5 \pm 6.0$ \\
%    & loss &  $0.428 \pm 0.017$  &  $0.584 \pm 0.122$  &  $0.337 \pm 0.096$  &  $0.366 \pm 0.082$  &  $0.584 \pm 0.122$ \\ 
%\hline
  \end{tabular}
  \label{tab:fmnist}
\end{table*}
% Domain weights: $0.350 \pm 0.001$, $0.316 \pm 0.002$, $0.334 \pm 0.002$, respectively.

\begin{table}[t]
    \centering
    \caption{Test perplexity of the train model on various domains, as
      a function of training loss for the language model dataset. \arxiv{Of all the model, the domain agnostic
      model that minimizes $\sL_{\sD_{\Lambda}}$ admits the best perplexity
      on the worst domain.}}
\vskip .15in
    \begin{tabular}{l c c c c} % \hline
    Training loss &  $\sU$  & \texttt{doc\icml{.}\arxiv{ument}} & \texttt{con\icml{.}\arxiv{versation}} & $\sD_{\Lambda}$ \\ \hline
    $\sL_{\texttt{doc\icml{.}\arxiv{ument}}}$ & $414.96$ & $83.97$ & $615.75$ & $615.75$ \\
    $\sL_{\texttt{con\icml{.}\arxiv{versation}}}$ & $108.97$ & $1138.76$ & $61.01$ & $1138.76$ \\
    $\sL_{\h \sU}$ & $68.18$ &  $96.98$ & $62.50$  & $96.98$ \\
    $\sL_{\sD_{\Lambda}}$ &  $79.98$ & $86.33$ & $78.48$ & $86.33$ \\
    \end{tabular}
    \label{tab:language}
\end{table}

To study the benefits of our AFL algorithm, we
carried out experiments with three datasets.  Even though our
optimization convergence guarantees hold only for convex
functions and stochastic gradient, we show that our domain-agnostic
learning performs well for non-convex functions and variants of
stochastic gradient descent such as momentum and Adagrad too.

In all the three experiments, we compare the domain agnostic model
with the model trained with $\h \sU$, the uniform distribution over
the union of samples, and the models trained on individual domains. In
all of these experiments, we used \textsc{PerDomain} stochastic
gradients and set $\Lambda = \Delta_p$. All algorithms were
implemented in Tensorflow \citep{AbadiEtAl2015}.

\subsection{Adult dataset}

The Adult dataset is a census dataset from the UCI Machine Learning
Repository \citep{blake1998uci}. \arxiv{It contains $32\mathord,561$
  training samples with numerical and categorical features, each
  representing a person.} The task consists of predicting if the
person's income exceeds $\$50\mathord,000$. We split this dataset into
two domains depending on whether the person had a doctorate degree or
not, resulting into domains: the \texttt{doctorate} domain
\arxiv{containing $413$ examples }and the \texttt{non-doctorate}
domain \arxiv{containing $32\mathord,148$ examples}. We trained a
logistic regression model with just the categorical features and
Adagrad optimizer. The performance of the models averaged over $50$
runs is reported in Table~\ref{tab:adult}. The performance on
$\sD_\Lambda$ of the model trained with $\h \sU$, that is standard
federated learning, is about $69.6\%$. In contrast, the performance of
our AFL model is at least about $71.5\%$ on \emph{any} target
distribution $\sD_\lambda$.  The uniform average over the domains of
the test accuracy of the AFL model is slightly less than that of the
uniform model, but the agnostic model is less biased and performs
better on $\sD_\Lambda$.  \arxiv{Furthermore, of the two domains, the
  \texttt{doctorate} domain is the harder one for predictions. For
  this domain, the performance of the domain agnostic model is close
  to the model trained only on \texttt{doctorate} data and is better
  than that of the model trained with the uniform distribution $\h
  \sU$.}

\subsection{Fashion MNIST}

The Fashion MNIST dataset, originally announced by \cite{xiao2017}, is
an MNIST-like dataset where images are classified into $10$ categories
of clothing, instead of handwritten digits. \arxiv{The dataset
  includes $60\mathord,000$ training images and $10\mathord,000$ test
  images given as 28x28 arrays of grayscale pixel intensities, spread
  evenly among the ten categories. We first trained a simple logistic
  regression classifier and observed that the lowest performance was
  achieved for the following three categories: \texttt{t-shirt/top},
  \texttt{pullover}, and \texttt{shirt}. Next, w}\icml{W}e extracted
the subset of the data labeled with \arxiv{these }three categories
\icml{\texttt{t-shirt/top}, \texttt{pullover}, and \texttt{shirt}} and
split this subset into three domains, each consisting of one class of
clothing. We then trained a classifier for the three classes using
logistic regression and the Adam optimizer. The results are shown in
Table~\ref{tab:fmnist}. Since here the domain uniquely identifies the
label, in this experiment, we did not compare against models trained
on specific domains. Of the three domains or classes, the
\texttt{shirt} class is the hardest one to distinguish from
others. The domain-agnostic model improves the performance for
\texttt{shirt} more than it degrades it on \texttt{pullover} and
\texttt{shirt}, leading to both \texttt{shirt}-specific and overall
accuracy improvement when compared to the model trained with the
uniform distribution $\h \sU$. Furthermore, in this experiment, note
that our agnostic learning solution not only improves the loss of the
worst domain, but also generalizes better and hence improves the
average test accuracy.  \arxiv{Our AFL model achieves a performance of
  about $\%74.5$ on \emph{any} target distribution $\sD_\lambda$,
  while the performance of standard federated learning can be as low
  as about $\%71.2$.}

\subsection{Language models}

Motivated by the keyboard application \citep{hard2018federated}, where
a single client uses a trained language model in multiple environments
such as chat apps, email, and web input, we created a dataset that
combines two very different types of language datasets:
\texttt{conversation} and \texttt{document}. For
\texttt{conversation}, we used the Cornell movie dataset that contain
movie dialogues \cite{danescu2011chameleons}. \arxiv{This dataset
  contains about $300\mathord,000$ sentences with an average sentence
  length of $8$.} For documents, we used the Penn TreeBank (PTB)
dataset\arxiv{ that contains approximately $50\mathord,000$ sentences
  with an average sentence length of $20$}
\cite{marcus1993building}. We created a single dataset by combining
both of the above corpuses, with \texttt{conversation} and
\texttt{document} as domains. We preprocessed the data to remove
punctuations, capitalized the data uniformly, and computed a
vocabulary of $10\mathord,000$ most frequent words. We trained a
two-layer LSTM model \arxiv{with LSTM and projection size of $512$}
with momentum optimizer. The performance of the models are measured by
their perplexity, that is the exponent of cross-entropy loss. The
results are reported in Table~\ref{tab:language}. Of the two domains,
the \texttt{document} domain is the one admitting the higher
perplexity. For this domain, the test perplexity of the domain
agnostic model is close to that of the model trained only on
\texttt{document} data and is better than that of the model trained
with the uniform distribution $\h \sU$.

\ignore{
\subsection{Online news popularity}

The Online News Popularity dataset of \cite{fernandes2015proactive},
found in the UCI Machine Learning Repository, provides attributes for
a set of 39,797 articles published by Mashable over a period of two
years. Six of these features indicate a category for the article; they
were removed as features and used as the domain, with the two smallest
("Lifestyle" and "Social Media") combined and a new "Other" domain for
those articles not in any of the original six categories. The number
of article shares on social media networks provides the target, with
1400+ considered "popular" and fewer than 1400 considered "not
popular." Data was randomly separated, with 80\% for training and 20\%
for testing. All features were normalized and fed into a DNN with two
hidden layers, both with ReLU activation functions, and trained using
the Adam optimizer with 20\% dropout on the input layer and 50\%
dropout on each hidden layer.

TODO(gsivek): results
}

\arxiv{\section{Extensions}
\label{sec:extensions}
In this section, we briefly discuss several extensions of the
framework, theory and algorithms that we presented.

\subsection{Domain definitions}

The choice of the domains can significantly impact learnability in
federated learning. In view of our learning bounds, if the number of
domains, $p$, is large and $\Lambda$ is the full simplex,
$\Lambda = \Delta_p$, then the models may not generalize well. Thus,
if the number of clients is very large, using each client as a domain
may be a poor choice for better generalization. Ideally, each domain
is represented with a sufficiently large number of samples and is
relatively homogeneous or pure. This suggests using a clustering
algorithm for defining the domains based on the similarity of the
client distributions. Different Bregman divergences could be used to
define the divergence or similarity between distributions.  Thus,
techniques such as those of \citet*{banerjee2005clustering} could be
used to determine clusters of clients using a suitable Bregman
divergence.

Client clusters can also be determined based on domain expertise. For
example, in federated keyboard next word prediction
\citep{hard2018federated}, domains can be chosen to be the native
language of the clients. If the model is used in variety of
applications, domains can also be based on the application of
interest. For example, the keyboard in \citep{hard2018federated} is
used in chat apps, social apps, and web inputs. Here, domains can be
the app that was used. Training models agnostically ensures that the
user experience is favorable in all apps.

\subsection{Incorporating a prior on $\Lambda$}

Agnostic federated learning as defined in \eqref{eq:AgnosticRisk}
treats all domains equally and does not incorporate any prior
knowledge of $\lambda$. Suppose we have a prior distribution
$p_{\Lambda}(\lambda)$ over $\lambda \in \Lambda$ at our disposal,
then, we can modify \eqref{eq:AgnosticRisk} to incorporate that
prior. If the loss function $\ell$ is the cross-entropy loss, then the
agnostic loss can be modified as follows:
\begin{equation}
\label{eq:prior}
\max_{\lambda \in \Lambda}
\left(\sL_{D_\lambda}(h) + \log p_\Lambda(\lambda) \right).
\end{equation}
In this formulation, larger weights are assigned to more likely
domains. The generalization guarantees of Theorem~\ref{thm:radamacher}
can be appropriately modified to include these changes. Furthermore,
if the prior $p_\Lambda(\lambda)$ is a log-concave function of
$\lambda$, then the new objective is convex in $h$ and concave in
$\lambda$ and a slight modification of our proposed algorithm can be
used to determine the global minima.  We note that we could also adopt
a multiplicative formulation with the prior multiplying the loss,
instead of the additive one with the negative log of the probability
in Equation~\ref{eq:prior}.

\subsection{Domain features and personalization}

We studied agnostic federated learning, where we learn a model that
performs well on all domains. First, notice that we do not make any
assumption on the hypothesis set $\sH$ and the hypotheses can use the
domain $k$ as a feature. Such models could be useful for applications
where the target domain is known at inference time. Second, while the
paper deals with learning a centralized model, the resulting model
$h_{\sD_\Lambda}$ can be combined with a personalized model, on the
client's machine, to design better client-specific models. This can be
done for example by learning an appropriate mixture weight
$\alpha_k \in [0, 1]$ to use a mixture
$\alpha_k h_{\sD_\Lambda} + (1 - \alpha_k) h_k$ of the domain agnostic
centralized model $h_{\sD_\Lambda}$ and a client- or domain-specific
model $h_k$.
}

\arxiv{
\section{Conclusion}

We introduced a new framework of AFL for which
we presented a detailed theoretical analysis. We also gave an
algorithm for this problem benefiting from our theoretical analysis,
as well as a new stochastic optimization solution needed for
large-scale problems. Our experimental results suggest that our
solution can lead to significant benefits in practice.}

\arxiv{
\section{Acknowledgements}
We thank Shankar Kumar, Rajiv Mathews, and Brendan McMahan for helpful
comments and discussions.}
\newpage
\bibliography{fed}
\icml{\bibliographystyle{icml2019}}
\newpage
\onecolumn
\appendix
\icml{

\section{Proof of Proposition~\ref{pro:lower_bound}}
\label{app:lower_bound}

\section{Proof of Theorem~\ref{thm:radamacher}}
\label{app:radamacher}

\section{Proof of Lemma~\ref{lem:vc}}
\label{app:vc}
  For any $\lambda \in \Lambda$, define the set of vectors $A_\lambda$
  in $\Rset^{m}$ by
\[
A_\lambda = \set[\bigg]{\bigg[\frac{\lambda_k}{m_k} \ell(h(x_{k, i}), y_{k,
      i}) \bigg]_{(k, i) \in [p] \times [m_k]} \colon \bx \in \sX^{m}, \by
    \in \sY^{m}}. 
\]
For any $\ba \in A_\lambda$, $\| \ba \|_2 = \sqrt{\sum_{k =
      1}^p m_k \frac{\lambda_k^2}{m_k^2}} = \sqrt{\sum_{k =
      1}^p \frac{\lambda_k^2}{m_k}} \leq
  \sqrt{\frac{\s(\Lambda \para \ov \bm)}{m}}$.
Then, by Massart's lemma, for any $\lambda \in \Lambda$, the following
inequalities hold:
\begin{align*}
\R_{\bm}(\sG, \lambda) 
& = \E_{\substack{S_k \sim \sD_k^{m_k} \\ \bsigma}} \left[\sup_{h \in \sH}
  \sum_{k = 1}^p \frac{\lambda_k}{m_k} \sum_{i = 1}^{m_k} \sigma_{k, i} \,
\ell(h(x_{k, i}), y_{k, i}) \right]\\
& \leq \E_{\bsigma} \left[\sup_{\ba
  \in A}
  \sum_{k = 1}^p \sum_{i = 1}^{m_k} \sigma_{k, i}a_{k, i} \right] \\
& \leq \sqrt{\frac{\s(\Lambda \para \ov \bm)}{m}} \,
  \frac{\sqrt{2 \log |A_\lambda|}}{m}\\
& = \frac{\sqrt{2 \s(\Lambda \para \ov \bm) \log |A_\lambda|}}{m}.
\end{align*}
By Sauer's lemma, the following holds for $m \geq d$:
$|A_\lambda| \leq \left(\frac{em}{d} \right)^d$. Plugging in the
right-hand side in the inequality above completes the proof.

}

\section{Alternative learning guarantees}
\label{app:alternative}

An objective similar to that of AFL was considered in the context of
multiple source domain adaptation by \citet{liu2015multiple}. The
authors presented generalization bounds for a scenario where the
target is based on some specific mixture $\lambda$ of the source
domains.  Our theoretical results differ from those of this work in two
ways. First, our generalization bounds do not hold for a single
mixture weight $\lambda$ but for any subset $\Lambda$ of the
simplex. Second, the complexity terms in the bounds presented by these
authors are proportional to
$\sqrt{m} \max_{k \in [p]} \frac{\lambda_k}{m_k}$, while our
guarantees are in terms of
$\sqrt{\sum_{k = 1}^p \frac{\lambda^2_k}{m_k}}$, which is strictly
tighter. In particular, in the special case where $k = 2$,
$\lambda_1 = \frac{1}{\sqrt{m}}$, $\lambda_2 = 1 - \lambda_1$ and
$m_1 = 1$ and $m_2 = m - 1$, the bounds of \citet{liu2015multiple} are
proportional to a constant and thus not informative,
$\sqrt{m} \max_{k \in [p]} \frac{\lambda_k}{m_k} = 1$, while our
guarantees are in terms of $\frac{1}{\sqrt{m}}$.

Our generalization error in Theorem~\ref{thm:radamacher} is
particularly useful when $\Lambda$ is a strict subset of the simple,
$\Lambda \subset \Delta_p$. If $\Lambda = \Delta_p$, we can give the
following alternative learning guarantee based. 
\begin{theorem}
\label{thm:dom_radamacher}
For any $\delta > 0$, with
probability at least $1 - \delta$ over the draw of samples
$S_k \sim \sD_k^{m_k}$, the following inequality holds for all
$h \in \sH$ and $\lambda \in \Lambda$:
\[
L_{\sD_\lambda}(h)   
\leq L_{\ov \sD_{\lambda}}(h) +
\sum^p_{k=1} \left( 2\lambda_k \R^k_{m_k}(\sG) + 
\lambda_k M \sqrt{\frac{1
}{2 m_k} \log \frac{p}{\delta}} \right),
\]
where $ \R^k_{m_k}(\sG) $ is the Rademacher complexity over domain
$\sD_k$ with $m_k$ samples.
\end{theorem}
The proof is a direct application of known Rademacher complexity
bounds \citep{MohriRostamizadehTalwalkar2012} and a union bound and is
omitted.

To relate the generalization bounds of Theorem~\ref{thm:radamacher}
and Theorem~\ref{thm:dom_radamacher}, observe that, by the
sub-additivity of $\sup$ and the linearity of expectation, the
following inequality holds:
\begin{align*}
\R_{\bm}(\sG, \lambda) 
& = \E_{\substack{S_k \sim \sD_k^{m_k} \\ \bsigma}} \left[\sup_{h \in \sH}
  \sum_{k = 1}^p \frac{\lambda_k}{m_k} \sum_{i = 1}^{m_k} \sigma_{k, i} \,
\ell(h(x_{k, i}), y_{k, i}) \right]\\ 
& \leq   \sum_{k = 1}^p \frac{\lambda_k}{m_k} 
\E_{\substack{S_k \sim \sD_k^{m_k} \\ \bsigma}} \left[\sup_{h \in \sH}
\sum_{i = 1}^{m_k}  \sigma_{k, i} \,
\ell(h(x_{k, i}), y_{k, i}) \right]\\ 
& = \sum^p_{k=1} \lambda_k \R^k_{m_k}(\sG).
\end{align*}
Furthermore, by the sub-additivity of $\sqrt{\cdot}$, the following
inequality holds:
\[
\sqrt{\frac{\s(\lambda \para \ov \bm)}{m}} = \sqrt{\sum^p_{k = 1}
  \frac{\lambda^2_k}{m_k}} 
\leq \sum^p_{k = 1} \sqrt{\frac{\lambda_k^2}{m_k}} 
= \sum^p_{k = 1} \lambda_k \sqrt{\frac{1}{m_k}}.
\]
Hence, up to the logarithmic factors in the second term, the guarantee
of Theorem~\ref{thm:radamacher} is stronger than that of
Theorem~\ref{thm:dom_radamacher}. However, $\Lambda_\epsilon$ can be
large and exponential in $p$, and it is not clear which of the bounds
are stronger in general. This depends on $\overline \bm$ and
$\lambda$. Deriving learning bounds that improve upon both of the
learning bounds above remains an interesting open question.

\icml{
\section{Proof of Theorem~\ref{thm:convergence}}
\label{app:convergence}

\section{Bounds on variaces of stochastic gradients}
\subsection{Proof of Lemma~\ref{lem:stochastic_lambda}}
\label{app:stochastic_lambda}
  The unbiasedness of $\delta_\lambda \L(w, \lambda)$ follows directly
  its definition. For the variance, observe that, for index
  $k \in [p]$, since the probability of not drawing domain $k$ is $(1
  - \frac{1}{p})$, the variance is given by the following
\begin{align*}
\var_k[\delta_\lambda \L(w, \lambda)]
& = \bigg[ 1 - \frac{1}{p} \bigg] [0 - \L_k(w)]^2 + \frac{1}{p} \sum_{k = 1}^p \frac{1}{m_k} \sum_{i = 1}^{m_k} \left[ p \L_{k,
  i}(w) - \L_k(w)  \right]^2 \\
& \leq \bigg[ 1 - \frac{1}{p} \bigg] M^2 + \frac{1}{p} \sum_{k = 1}^p
  \frac{1}{m_k} \sum_{i = 1}^{m_k} [pM]^2 = pM^2.
\end{align*}
Summing over all indices from $k \in [p]$ completes the proof.

\subsection{Proof of Lemma~\ref{lem:perdomain}}
\label{app:perdomain}
  The time complexity and the unbiasedness follow from the
  definitions. We now bound the variance.  Since $\nabla_w \L_{k, J_k}$
  is an unbiased estimate of $\nabla_w \L_k(w)$ and we have:
\begin{align*}
\var[\delta_w] = \sum_{k = 1}^p \lambda^2_k  \var\left[
\nabla_w \L_{k, J_k}(w) - \nabla_w \L_k(w)  \right] 
\leq \sum_{k = 1}^p \lambda^2_k  \sigma^2(w, I) 
\leq R_{\Lambda} \sigma^2_I(w).
\end{align*}
This completes the proof.

\subsection{Proof of Lemma~\ref{lem:weighted}}
\label{app:weighted}
  The time complexity and the unbiasedness follow from the
  definitions. We now bound the variance. By definition for any
  $w, \lambda$,
\begin{align*}
\var(\delta_w) &
 = \sum^p_{k = 1} \frac{\lambda_k}{m_k} \sum^{m_k}_{j = 1} 
 \left( \nabla_w \L_{k, j}(w) - \L(w, \lambda) \right)^2 \\
& = \sum^p_{k = 1} \frac{\lambda_k}{m_k} \sum^{m_k}_{j = 1} 
 \left( \nabla_w \L_{k, j}(w) - \L_k(w) \right)^2 + 
 \sum^p_{k = 1} \lambda_k (\L_k(w) - \L(w, h))^2 \\
& \leq \sigma^2_I(w) + \sigma^2_O(w),
\end{align*}
where the second equality follows from the unbiasedness of the stochastic gradients.

}

\end{document}